\documentclass[11pt,twocolumn,letterpaper]{article}
\usepackage[margin=1in]{geometry}
\usepackage{amssymb}
\usepackage{amsmath,amsfonts,bm}

\def\eqref#1{equation~\ref{#1}}
\def\1{\bm{1}}

\DeclareMathAlphabet{\mathsfit}{\encodingdefault}{\sfdefault}{m}{sl}
\SetMathAlphabet{\mathsfit}{bold}{\encodingdefault}{\sfdefault}{bx}{n}

\usepackage{graphicx,wrapfig}
\usepackage{booktabs,tabularx}
\usepackage[table]{xcolor}
\usepackage{algorithm,algpseudocode}
\usepackage[numbers,sort&compress]{natbib}
\usepackage[hidelinks]{hyperref}

\makeatletter
\providecommand{\@LN@col}[1]{}
\providecommand{\@LN}[2]{}
\makeatother

\title{Let the Mind Conceive, Let the Hand Achieve: Coupling Grounded Action and Progress Evaluation for Long-Horizon Manipulation}
\author{Youpeng Wen$^{1}$, Gang Wang$^{1}$, Tianyu Huang$^{1}$, Weiwei Zhang$^{1}$,\\
Zhijian Huang$^{2}$, Wei Chen$^{1}$, Xieyuanli Chen$^{3}$, Yunhui Liu$^{1}$\\[0.5em]
\normalfont $^{1}$ The Chinese University of Hong Kong\\
\normalfont $^{2}$ Sun Yat-sen University\\
\normalfont $^{3}$ National University of Defense Technology}
\date{}

\begin{document}
\twocolumn[{
\begin{@twocolumnfalse}
\maketitle
\begin{abstract}
A robot following a multi-step instruction needs to act on the right objects in the right order, even as the scene changes. It must also maintain awareness of task progress and determine when to advance.
To address these challenges, we introduce Orchestra, an agent harness that separates task planning and execution management from action generation.
Orchestra decomposes a complete multimodal instruction into ordered subtasks, tracks completed steps, and updates the locations of relevant objects and destinations during execution.
It maintains a shared context containing the active subtask, spatial grounding, and recent observations.
An action policy uses this context to generate actions, while a separately trained progress evaluator estimates when to advance to the next subtask.
To evaluate whether robots trained on atomic instructions and short trajectories can execute longer subtask sequences, we introduce a benchmark named RMBench-Compose.
Orchestra achieves a mean success rate of $68.22\%$ across nine RMBench tasks, outperforming Mem-0 by $26.22$ percentage points.
On RMBench-Compose, it achieves an Average Success Length (ASL) of $7.35$ subtasks, $2.46$ times that of HarnessVLA.
Across four real-world multi-step tasks, it improves mean ASL by $25.0\%$ over Orchestra using the base $\pi_{0.5}$ policy.
These results show that Orchestra helps robots carry out longer sequences of subtasks while preserving the requested subtask order and acting on the specified objects.

Our code and project page are available at \url{https://youpengwen.github.io/Orchestra}.
\end{abstract}

\end{@twocolumnfalse}
}]
\section{Introduction}
\label{sec:introduction}

% Preamble:
% \usepackage{graphicx}

\begin{figure*}[t]
    \centering
    \includegraphics[width=\textwidth]{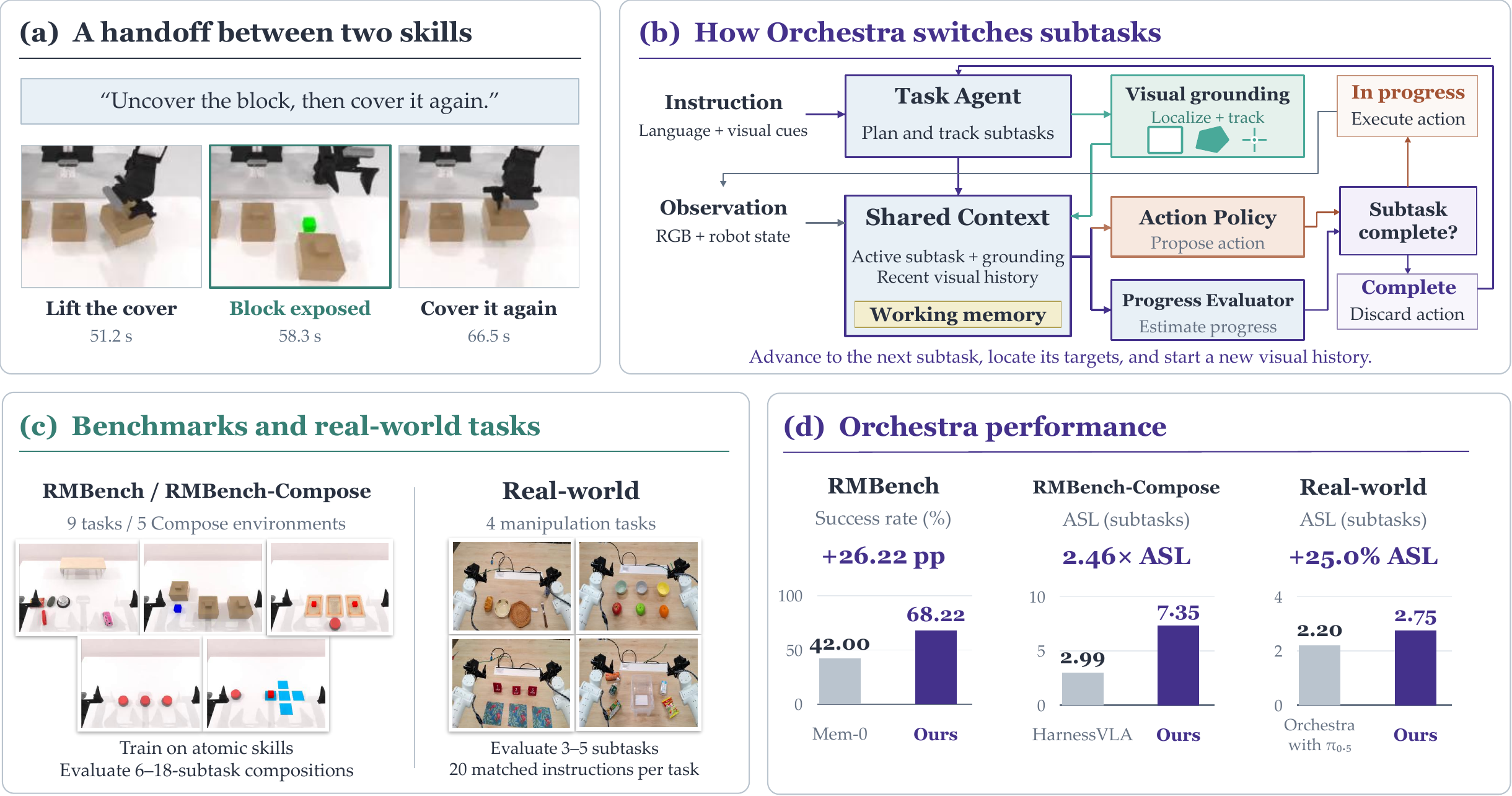}
    % \caption{
    % \textbf{Orchestra for long-horizon instruction following.}
    % \textbf{(a)} Executing a complete instruction requires maintaining
    % the requested subtask order and grounding each operation in the
    % intended objects and destinations.
    % \textbf{(b)} Orchestra maintains a shared execution context that
    % aligns the active subtask, spatial grounding, and recent visual
    % history for action generation and progress-driven subtask transitions.
    % \textbf{(c)} We evaluate Orchestra on nine RMBench tasks,
    % five RMBench-Compose environments, and four real-world tasks.
    % RMBench-Compose evaluates sequences of 6--18 subtasks after training
    % on atomic skills; real-world evaluation uses 20 matched instructions
    % per task.
    % \textbf{(d)} Orchestra improves mean RMBench success rate by
    % $26.22$ percentage points over Mem-0, achieves $2.46\times$ the
    % mean Average Success Length (ASL) of HarnessVLA on RMBench-Compose,
    % and achieves $1.25\times$ the mean ASL of Orchestra with
    % $\pi_{0.5}$ on real-world tasks.
    % ASL measures consecutive successful subtasks, not full-task success.
    % Both real-world configurations use the Orchestra harness.
    % Results are unweighted task averages without uncertainty intervals.
    % }
    \vspace{-4mm}
    \caption{
\textbf{How Orchestra hands control from one skill to the next.}
\textbf{(a)} After uncovering the green block, the robot must find the displaced cover to put it back.
\textbf{(b)} Orchestra executes the proposed action while the subtask is in progress. On completion, it discards the proposal, grounds the next subtask in the current scene, and starts a new visual history.
\textbf{(c)} We evaluate Orchestra on RMBench, RMBench-Compose, and four real-world tasks.
\textbf{(d)} On RMBench-Compose, Orchestra completes 7.35 consecutive subtasks on average, compared with 2.99 for HarnessVLA.
}
    \label{fig:teaser}
    \vspace{-5mm}
\end{figure*}

Consider the instruction \emph{Uncover the block, then cover it again.} In the middle frame of Figure~\ref{fig:teaser}(a), the green block is exposed and its cover rests just in front of it. This scene is the goal of the first skill and the starting point of the second.
The instruction gives the order in advance. The robot must still recognize when it has finished uncovering the block and find the cover where it set it down.
How should a robot trained to execute individual skills determine which part of a longer instruction should govern its next action?

% p2
This difficulty is visible even when the individual skills are familiar. On \textbf{RMBench-Compose}, our benchmark built on five RMBench environments~\citep{chen2026rmbench}, a base $\pi_{0.5}$ policy~\citep{physicalintelligence2025pi05} succeeds on $85.0\%$ of isolated atomic trials.
Given a single instruction specifying 6--18 subtasks, the same policy completes only $0.93$ consecutive subtasks on average.
% For execution from a complete instruction, perceptual history and task records must remain aligned with the active subtask.
Generalist vision-language-action (VLA) policies provide a repertoire of language-conditioned skills~\citep{brohan2023rt2,octo2024,kim2024openvla,physicalintelligence2025pi05}. MEM gives policies access to recent observations and task-level memory~\citep{torne2026mem}, and HarnessVLA uses an agent to compose learned and analytic primitives~\citep{zhang2026harnessvla}.
We study the handoff between two skills. When one skill ends, what information should carry forward, and what must be established anew for the next?

% p3
We introduce \textbf{Orchestra}, which interprets a complete multimodal instruction once and determines during execution when each subtask takes control (Figure~\ref{fig:teaser}).
The plan can specify in advance \emph{which} object a later step will need. Its location when that step begins depends on what the robot has actually done.
Orchestra retains the ordered plan and a record of completed steps throughout the episode. As each subtask becomes active, it locates the relevant objects and destinations in the current scene and begins a visual history for that subtask.

% p4
% Language-model agents offer a useful perspective on organizing these capabilities.
% ReAct interleaves reasoning with actions and observations~\citep{yao2023react}.
% In robotics, SayCan grounds skill selection in learned affordances~\citep{ahn2022saycan}, Inner Monologue incorporates execution feedback into planning~\citep{huang2022innermonologue}, and Code as Policies composes perception and control interfaces through generated programs~\citep{liang2022codeaspolicies}.
% More recently, HarnessVLA uses a memory-guided agent to compose a frozen VLA for contact-rich interactions with analytic manipulation primitives~\citep{zhang2026harnessvla}.
% These systems motivate explicit coordination around learned action capabilities.
% Our focus is an agent harness that maintains a shared execution context, aligning the active subtask, spatial grounding, observation history, and progress-driven transitions throughout a complete instruction.

% p4
% Language-model agents coordinate reasoning, action, and feedback~\citep{yao2023react}, with robotic systems incorporating affordances, execution feedback, and programmatic control~\citep{ahn2022saycan,huang2022innermonologue,liang2022codeaspolicies}.
% HarnessVLA further uses a memory-guided agent to compose a frozen VLA with analytic manipulation primitives~\citep{zhang2026harnessvla}.
The handoff takes place inside the action loop. At each replanning call, the action policy proposes a short sequence of movements while a separately trained evaluator estimates progress on the same grounded subtask.
If the evaluator predicts completion, Orchestra discards the proposed movements and prepares the next skill. In Figure~\ref{fig:teaser}(a), recent observations of the cover moving away from the green block help the evaluator recognize when uncovering is complete. Orchestra then locates the displaced cover and the exposed block so that the next action can put the cover back.
The same decision thus changes both the instruction that governs the next action and the visual evidence supplied with it.

On RMBench-Compose, the action policy is fine-tuned on atomic instructions and short demonstrations, and each test episode begins with one instruction for the entire sequence.
Holding the base $\pi_{0.5}$ action policy fixed, Orchestra increases Average Success Length (ASL), the mean number of consecutive subtasks completed, from $0.93$ to $3.60$. Conditioning the action policy on spatial prompts and recent visual history raises ASL to $7.35$, compared with $2.99$ for HarnessVLA.

% % p7
% We evaluate Orchestra on \textbf{RMBench}~\citep{chen2026rmbench}, \textbf{RMBench-Compose}, and \textbf{four real-world manipulation tasks}.
% We introduce RMBench-Compose by adapting five RMBench environments to support variable-length, state-feasible compositions of atomic skills.
We also evaluate the handoff directly. Across 500 cases per configuration, increasing the evaluator's visual input from one frame to six raises its on-time switch rate from $77.0\%$ to $96.6\%$. On four physical-robot tasks, the conditioned action policy increases mean ASL from $2.20$ to $2.75$ within the same Orchestra system.

% Our contributions are threefold:
% \begin{itemize}
%     \item We introduce \textbf{Orchestra}, an agent harness that coordinates task interpretation, online spatial grounding, working memory, action generation, and progress-driven subtask transitions through a shared structured context.
    
%     \item We develop context interfaces that provide task-relevant spatial conditions and short-horizon visual history to an action policy and an independent progress evaluator, with explicit context maintenance across subtask boundaries.
    
%     \item We introduce \textbf{RMBench-Compose} for compositional execution from complete instructions and evaluate Orchestra on RMBench, RMBench-Compose, and four real-world manipulation tasks.
% \end{itemize}

\section{Related Work}

\textbf{Imitation-based robot policies.} Recent robot learning has increasingly focused on building generalist manipulation policies from large-scale and heterogeneous demonstrations~\citep{kim2024openvla,black2025pi0,physicalintelligence2025pi05}. Early vision-language-action (VLA) models transfer semantic representations from pre-trained vision-language models to robot control; they typically discretize continuous actions and predict them as tokens~\citep{brohan2023rt2,kim2024openvla}. More recent approaches instead condition dedicated action experts on vision-language representations, enabling more expressive continuous control~\citep{li2024cogact}. For example, $\pi_0$ uses a flow-matching action expert to generate temporally chunked actions~\citep{black2025pi0}, while $\pi_{0.5}$ combines heterogeneous co-training with semantic subtask prediction for open-world generalization~\citep{physicalintelligence2025pi05}. These models mainly improve manipulation capability through policy scaling, broader training data~\citep{walke2023bridgedata}, and stronger action representations. In this work, we take a complementary approach: We retain the pretrained VLA policy as the action policy and externalize the planning, memory, and command-mapping functions to the agent-level system.

\textbf{Long-horizon manipulation.} Long-horizon manipulation requires tracking task progress~\citep{kang2025taksie,liu2026palm,huang2026stageguard} and relevant history beyond the current observation. Recent approaches address this through hierarchical task decomposition~\citep{zhu2022buds,zhang2024uvd} and temporal memory. Hierarchical methods separate coarse task semantics from low-level control~\citep{shiarlis2018taco,belkhale2024rth}; for example, MimicPlay uses high-level guidance derived from human play~\citep{wang2023mimicplay}, while $\pi_{0.5}$ predicts semantic subtasks that condition subsequent action generation~\citep{physicalintelligence2025pi05}. More recent work  incorporates historical context into VLA policies~\citep{shi2025memoryvla}. HAMLET aggregates compact moment tokens through a lightweight memory module, allowing pretrained VLAs to efficiently use past observations~\citep{koo2026hamlet}. MEM further introduces memory at multiple temporal scales, combining short-term perceptual context with long-term semantic task information~\citep{torne2026mem}. RMBench conducted a systematic study of this configuration and proposed Mem-0 to enable memory-based operations using explicit memory components~\citep{chen2026rmbench}. These methods primarily incorporate temporal context into the policy or its inference architecture. In contrast, our framework combines policy-level visual memory with explicit task records and subtask-transition control maintained by the agent harness.

\textbf{Robot agents.} Recent robot-agent systems use foundation models to coordinate reusable robot skills, separating high-level reasoning from low-level execution~\citep{ahn2022saycan,huang2022innermonologue}. Early approaches such as Code as Policies generate executable programs over perception and control APIs, providing a modular interface between language-model reasoning and robot control~\citep{liang2022codeaspolicies}. Later systems extend this idea to large-scale embodied orchestration. AutoRT combines VLM-based scene understanding with LLM-based task generation to coordinate robot behaviors in real-world environments~\citep{ahn2024autort}. More recent work integrates VLAs directly into the agent hierarchy. VLAs-as-Tools treats specialized VLA policies as executable tools selected by a high-level VLM agent. It also uses execution feedback and event-triggered replanning to support long-horizon tasks~\citep{lei2026toolaligned}. HarnessVLA takes a different approach. It uses a frozen VLA as a contact-rich manipulation primitive and coordinates it with analytic primitives through a memory-guided agent~\citep{zhang2026harnessvla}. These methods shift robot agents from predefined symbolic skills toward learned VLA policies. Our framework follows this direction, emphasizing external task coordination, multi-timescale memory, and multimodal instruction anchoring around a pretrained VLA.

\begin{figure*}[ht]
\centering
\includegraphics[width=\textwidth]{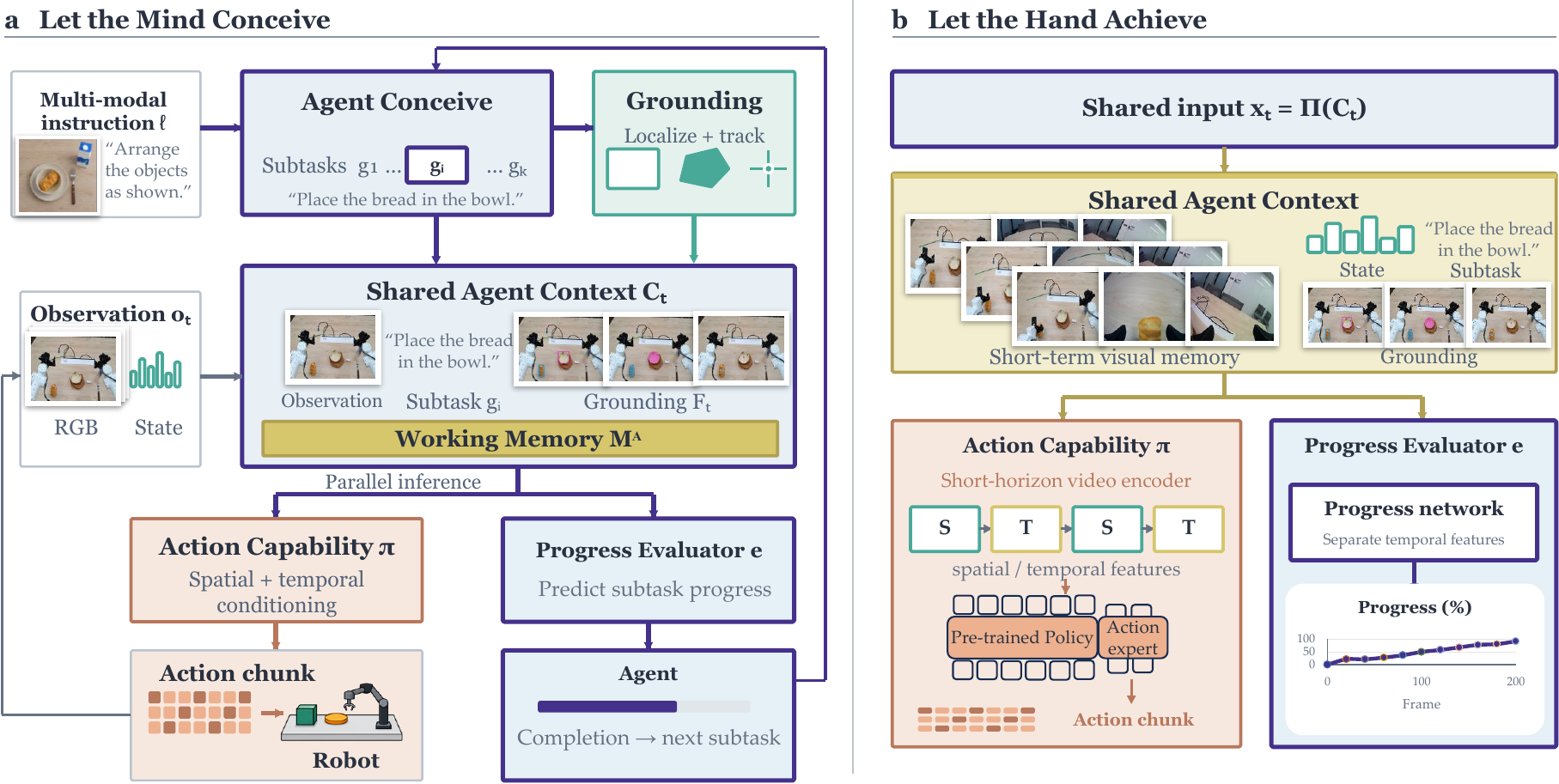}
% \vsapce{-3mm}
\caption{\textbf{Orchestra framework.} \textbf{(a) Let the Mind Conceive.} The agent decomposes multimodal instructions into subtasks, grounds relevant entities, and maintains task records and recent RGB observations. \textbf{(b) Let the Hand Achieve.} An action policy and an independent progress evaluator use the active subtask, observations, and spatial grounding to generate action chunks and predict progress. A terminal-category prediction triggers the next subtask. }
\label{fig:orchestra-framework}
\vspace{-3mm}
\end{figure*}

% Compressed alternative to 4_method.tex; include instead of the original.
% Required packages: amsmath, amssymb, graphicx, hyperref.
% Configuration: K=6, Delta=3, B=11; report in Experimental Setup.

\section{Method}
\label{sec:orchestra-method}

\subsection{From a Complete Instruction to the Current Skill}

An instruction can identify an object long before the robot needs to act on it. A request to arrange numbered plates in the order \(3,2,1\) fixes the identity of the second plate at the start of the episode. The robot needs its location only after it has placed plate~3. Orchestra therefore determines the order and entity bindings from the complete instruction, then locates each skill's referents in the scene in which that skill begins (Figure~\ref{fig:orchestra-framework}).

\textbf{Task interpretation.} The input \(\ell\) can be a language instruction, language paired with a reference image~\citep{jiang2022vima}, or a referring instruction accompanied by user-provided points or bounding boxes. The orchestration agent resolves these references into an ordered plan \(\mathcal G=(g_1,\ldots,g_N)\), where each \(g_i\) specifies an executable instruction and the relevant objects or destinations. We use DeepSeek-V4.1-Flash~\citep{deepseekai2026v41flash} and evaluate Qwen3.7-Plus~\citep{qwen2026qwen37plus} as an alternative (Section~\ref{sec:results-compose}). The action policy receives the active instruction and its scene grounding.

\textbf{Subtask-conditioned grounding.} MolmoPoint-8B~\citep{clark2026molmopoint} locates task-relevant entities in the current head-camera image, providing point prompts for SAM3~\citep{carion2026sam3} instance segmentation. Grounding is initialized at subtask boundaries; SAM3 updates the selected regions at subsequent action replanning calls. The resulting masks provide bounding boxes, sampled points, or segmentation contours for downstream conditioning.

\subsection{Acting and Judging Progress Within a Skill}
\label{sec:orchestra-policy}

At each replanning call, two questions concern the active skill. What movement would continue it? Does it still need to continue? Orchestra computes both predictions before executing the proposed movement. The action policy \(\pi_\theta\) proposes a chunk \(A_t\), and a separately trained evaluator \(e_\phi\) predicts a distribution \(q_t\) over \(B\) within-skill progress categories:
\begin{equation}
\label{eq:orchestra-1}
\begin{gathered}
C_t=\bigl(o_t,g_{i_t},M_t^A,F_t\bigr),\\
A_t=\pi_\theta(C_t),\qquad q_t=e_\phi(C_t),
\end{gathered}
\end{equation}
Here \(o_t=(\mathbf I_t,s_t)\) contains multi-camera images and proprioception, \(g_{i_t}\) is the active instruction, and \(F_t\) is its spatial grounding. The agent maintains the working memory
\begin{equation}
\label{eq:orchestra-2}
M_t^A=(\mathcal D_t,\mathcal H_t,i_t),
\end{equation}
where \(\mathcal D_t\) records completed steps, \(\mathcal H_t\) is a bounded history of timestamped multi-camera observations from the active skill, and \(i_t\) is the plan index. Each model's adapter selects its inputs from \(C_t\). Both predictions refer to the same target and recent scene history.

\textbf{Action prediction.} We adapt \(\pi_{0.5}\)~\citep{physicalintelligence2025pi05} to generate actions from the active instruction, spatial prompts, proprioception, and recent RGB observations. Its PaliGemma~\citep{beyer2024paligemmaversatile3bvlm} backbone uses factorized spatial and temporal attention before the action expert predicts the chunk.

\textbf{Spatial conditioning.} During training, each demonstration is paired with spatial annotations for entities relevant to the active subtask. Bounding boxes and points are serialized as entity names with quantized coordinates, e.g., \texttt{<bbox, "brown loaf", [x1,y1,x2,y2]>} or \texttt{<point, "brown loaf", [x,y]>}, and appended to the language prefix alongside the subtask instruction. For mask-based conditioning, segmentation contours are overlaid on the corresponding objects in the head-camera image. 
%Both the action policy and progress evaluator receive these spatial conditions together with the subtask instruction, proprioception, and visual history.

\textbf{Temporal visual encoding.} For each camera \(c\), we sample a window from \(\mathcal H_t\):
\begin{equation}
\label{eq:orchestra-3}
\mathcal W_t^c=\left(I_{t-(K-1)\Delta}^{c},\ldots,I_{t-\Delta}^{c},I_t^{c}\right),
\end{equation}
where \(K\) is the window length and \(\Delta\) the sampling stride; unavailable history is masked. Spatial attention operates within frames, and temporal attention mixes features at corresponding patch positions across frames~\citep{bertasius2021timesformer,torne2026mem}. Only current-frame features are retained for the language-model prefix after temporal fusion. History therefore adds temporal computation without increasing the number of visual tokens passed to the language model.

\textbf{Action learning.} The policy predicts action chunks of length \(H_p\) using the flow-matching objective of the base policy~\citep{lipman2023flowmatching}, conditioned on the subtask instruction, spatial prompts, and recent visual observations.

\textbf{Progress prediction.} An independent evaluator uses the \texttt{gemma\_300m} configuration from OpenPI~\citep{physicalintelligence2025pi05} for its language backbone. It receives the active subtask, spatial conditions, proprioception, and recent visual history. A learnable progress token aggregates these inputs into \(h_t\), yielding
\begin{equation}
\label{eq:orchestra-5}
q_t=\operatorname{softmax}\!\left(W_{\mathrm{prog}}\operatorname{LN}(h_t)+b_{\mathrm{prog}}\right).
\end{equation}
The predicted category is \(\hat b_t=\arg\max_{j\in\{0,\ldots,B-1\}}q_{t,j}\).

\textbf{Progress supervision.} Demonstrations are segmented into language-labeled subtasks~\citep{chen2026sarm}. For frame \(u\) in segment \([a_i,e_i)\), temporal progress and its discrete label are
\begin{equation}
\label{eq:orchestra-6}
\begin{gathered}
\rho_i(u)=1-\frac{\max(e_i-1-u,0)}{\max(e_i-a_i,1)},\\
b_i(u)=\left\lceil B\rho_i(u)\right\rceil-1.
\end{gathered}
\end{equation}
We train the evaluator with cross-entropy:
\begin{equation}
\label{eq:orchestra-7}
\mathcal L_{\mathrm{prog}}=-\mathbb E_{(u,i)}\log q_{u,b_i(u)}.
\end{equation}
The terminal label covers the final interval \((1-1/B,1]\) of normalized demonstration progress.

\subsection{What Carries Across a Skill Boundary?}

A terminal progress prediction changes what happens to the action already waiting to be executed. At each replanning call, the agent updates grounding, queries the policy and evaluator in parallel, and computes the switch decision
\begin{equation}
\label{eq:orchestra-8}
c_t=\mathbf 1[\hat b_t=B-1].
\end{equation}
When \(c_t=0\), the agent executes \(A_t\). Between replanning calls, intermediate observations update the visual history. When \(c_t=1\), the agent discards \(A_t\), records \(g_{i_t}\) as complete, and advances the plan.

The completed-step record persists across the handoff. If a skill remains, Orchestra starts a fresh visual history and grounds the new skill's referents in the current scene before requesting its first action. The last scene of one skill thus becomes the starting scene of the next, now interpreted under a different instruction. Advancing beyond the final skill ends the plan. Task success is measured independently of the agent's decision to end the plan.

\section{RMBench-Compose Benchmark}
\label{sec:compositional-benchmark}

% \subsection{Task Construction and Evaluation}

\begin{table*}[t]
\centering
\caption{\textbf{Success rates on RMBench.} 
% Total average is the unweighted mean across nine tasks. 
The best result in each column is shown in bold.}
\label{tab:orchestra-rmbench}
\begingroup
\definecolor{OrchestraInk}{HTML}{35258B}
\setlength{\tabcolsep}{3.5pt}
\setlength{\aboverulesep}{1pt}
\setlength{\belowrulesep}{1pt}
\renewcommand{\arraystretch}{1.1}
\footnotesize
\resizebox{\textwidth}{!}{%
\begin{tabular}{@{}l*{10}{c}@{}}
\toprule
\rowcolor{OrchestraInk!10}
\textbf{Method} & \shortstack{\textbf{Observe and}\\\textbf{Pick Up}} & \shortstack{\textbf{Rearrange}\\\textbf{Blocks}} & \shortstack{\textbf{Put Back}\\\textbf{Block}} & \shortstack{\textbf{Swap}\\\textbf{Blocks}} & \shortstack{\textbf{Swap}\\\textbf{T}} & \shortstack{\textbf{Battery}\\\textbf{Try}} & \shortstack{\textbf{Blocks}\\\textbf{Ranking Try}} & \shortstack{\textbf{Cover}\\\textbf{Blocks}} & \shortstack{\textbf{Press}\\\textbf{Button}} & \shortstack{\textbf{Total}\\\textbf{average}} \\
\midrule
\rowcolor{black!6}
\multicolumn{11}{l}{\textcolor{black!60}{\textit{Base policies}}} \\
DP~\citep{chi2023diffusionpolicy} & 1\% & 0\% & 0\% & 11\% & \textbf{20\%} & 10\% & 10\% & 0\% & 0\% & 5.78\% \\
ACT~\citep{zhao2023act} & 1\% & 29\% & 0\% & 2\% & 2\% & 19\% & 0\% & 0\% & 0\% & 5.89\% \\
$\pi_{0.5}$~\citep{physicalintelligence2025pi05} & 9\% & 13\% & 11\% & 24\% & 15\% & 16\% & 6\% & 0\% & 0\% & 10.44\% \\
X-VLA~\citep{zheng2025xvla} & 9\% & 13\% & 18\% & 16\% & 3\% & 26\% & 1\% & 2\% & 0\% & 9.78\% \\
QwenOFT~\citep{starvla2026} & 0\% & 0\% & 0\% & 0\% & 0\% & 14\% & 37\% & 0\% & 0\% & 5.67\% \\
\midrule[0.5pt]
\rowcolor{black!6}
\multicolumn{11}{l}{\textcolor{black!60}{\textit{Dual-system models}}} \\
MemER~\citep{sridhar2025memer} & 7\% & 17\% & 0\% & 14\% & 7\% & 27\% & 0\% & 6\% & 0\% & 8.67\% \\
Mem-0~\citep{chen2026rmbench} & 4\% & 89\% & \textbf{90\%} & 67\% & 14\% & 28\% & 18\% & 68\% & 0\% & 42.00\% \\
\midrule[0.5pt]
\rowcolor{black!6}
\multicolumn{11}{l}{\textcolor{black!60}{\textit{End-to-end memory models}}} \\
MemoryVLA (OpenVLA)~\citep{shi2025memoryvla} & 0\% & 22\% & 50\% & 17\% & 9\% & 25\% & 12\% & 40\% & 0\% & 19.44\% \\
MemoryVLA (QwenOFT)~\citep{shi2025memoryvla,yang2026eventvla} & 2\% & 53\% & 81\% & 76\% & 9\% & \textbf{33\%} & 53\% & \textbf{69\%} & 0\% & 41.78\% \\
\midrule[0.5pt]
\rowcolor{black!6}
\multicolumn{11}{l}{\textcolor{black!60}{\textit{Agent-based systems}}} \\
HarnessVLA~\citep{zhang2026harnessvla} & 10\% & 32\% & 62\% & 18\% & 5\% & 24\% & 20\% & 26\% & 44\% & 26.78\% \\
\midrule[0.5pt]
\rowcolor{OrchestraInk!10}
\textcolor{OrchestraInk}{\textbf{Orchestra}} & \textbf{81\%} & \textbf{95\%} & 81\% & \textbf{86\%} & 10\% & 30\% & \textbf{65\%} & \textbf{69\%} & \textbf{97\%} & \textbf{68.22\%} \\
\bottomrule
\end{tabular}%
}
\endgroup
\end{table*}

We introduce RMBench-Compose by adapting five RMBench~\citep{chen2026rmbench}
environments for long-horizon instruction following.
Whereas the original tasks use predefined instruction templates and
execution sequences, RMBench-Compose supports varying the length and
order of feasible operations within each environment.
The benchmark evaluates whether systems can follow complete language
instructions over sequences longer than those seen during atomic-skill
training.
Successful execution requires maintaining the requested operation order
and selecting the intended objects and destinations throughout the sequence.

The five environments are \emph{Press Button}, \emph{Pick Up Object},
\emph{Cover and Uncover Block}, \emph{Swap Block}, and
\emph{Place Block on Mat}.
Each environment defines atomic subtasks with associated language
instructions, relevant entities, feasibility conditions, and success predicates.
Feasibility depends on the current scene state; for example, an uncovering
operation requires the target block to be covered.

Let \(\mathcal U_e\) be the atomic subtask library for environment \(e\),
and \(\mathcal F_e(x)\subseteq\mathcal U_e\) its feasible subset at state \(x\).
A composed task is
\begin{equation}
\mathcal G=(g_1,\ldots,g_L),\qquad
g_j\in\mathcal F_e(x_{j-1}),
\label{eq:benchmark-composition}
\end{equation}
where \(x_{j-1}\) is the scene state after successful execution of the
preceding subtasks.
Sequence length and subtask order are configurable, subject to the
environment's available skills and state-dependent feasibility conditions.

\textbf{Evaluation protocol.}
Training demonstrations contain atomic instructions and short trajectories.
At evaluation time, each episode begins with a complete language
instruction \(\ell\) specifying the ordered subtask sequence.
Each subtask starts from the scene state left by the preceding subtask
and has an environment-specific action budget.
Episodes end when the sequence is completed or a subtask exhausts its
action budget.

We report Atomic Success Rate (Atomic SR) on isolated atomic tasks and
two metrics for composed-task execution.
Average Success Length (ASL) measures the mean number of consecutive
subtasks completed in the requested order~\citep{mees2022calvin}.
Subtask Success Rate (SSR) measures the fraction of attempted subtasks
completed successfully, excluding subtasks not reached after an earlier failure.

% Alternative experiment section; replace the original input rather than including both.
% Original source: files/6_experiment.tex. Figure and table assets are unchanged.

\section{Experiments}
\label{sec:experiments}

We evaluate sustained instruction following on RMBench~\citep{chen2026rmbench}, RMBench-Compose, and a physical robot, with ablations examining the components of Orchestra.

\subsection{Experimental Setup}
\label{sec:experiment-setup}

\textbf{Benchmarks and task inputs.}
We evaluate nine RMBench tasks. For the five RMBench-Compose environments, we use the atomic-training, composed-evaluation protocol in Section~\ref{sec:compositional-benchmark}.
Evaluation lengths are $n=10,12,6,14,18$ for Pick Up Object, Cover and Uncover Block, Press Button, Swap Block, and Place Block on Mat, respectively.
We conduct 100 evaluation rollouts for each task.
Training uses atomic instructions and short demonstrations, whereas evaluation begins with the complete ordered instruction and preserves scene states across subtask boundaries.
% These environment-specific lengths do not constitute a controlled length sweep.

\begin{figure*}[ht]
\centering
\includegraphics[width=\textwidth]{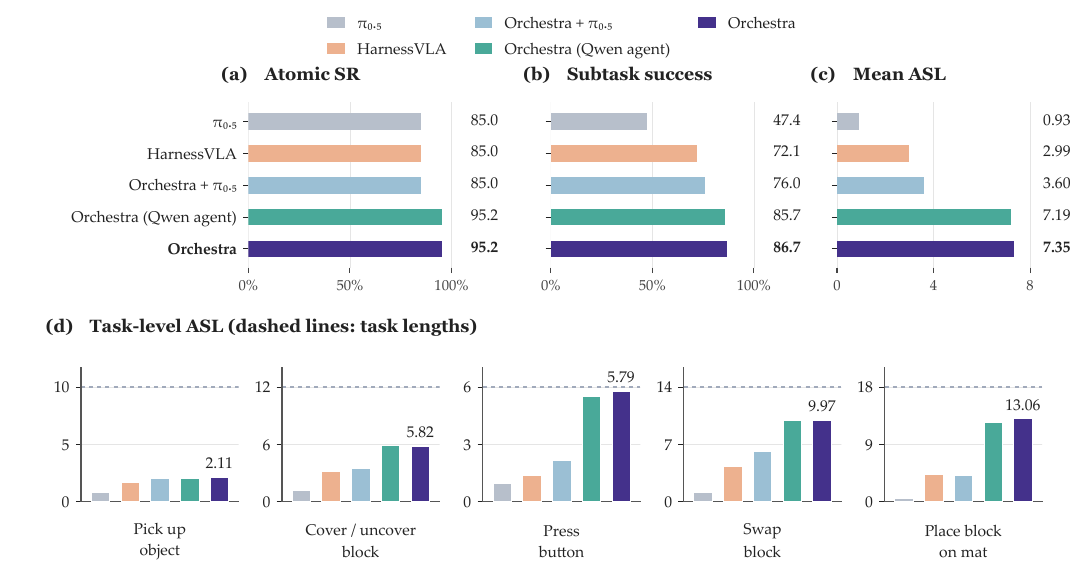}
\caption{\textbf{Atomic and compositional performance on RMBench-Compose.} \textbf{(a--c)} Unweighted five-environment averages of Atomic SR, SSR, and ASL. \textbf{(d)} Per-environment ASL; dashed segments indicate task lengths. Orchestra + $\pi_{0.5}$ uses the base action policy. Orchestra (Qwen agent) replaces the orchestration agent.}
\vspace{-3mm}
\label{fig:results-compose}
\end{figure*}

\textbf{Compared configurations.}
Table~\ref{tab:orchestra-rmbench} combines nine officially reported RMBench baselines~\citep{chen2026rmbench,yang2026eventvla} with our evaluations of HarnessVLA and Orchestra.
For our HarnessVLA reproduction~\citep{zhang2026harnessvla}, we train the $\pi_{0.5}$ policy on the same subtask demonstrations as Orchestra on each benchmark.
In the RMBench few-shot protocol, the Codex planner explores one reference seed per task with resets, saving the successful primitive sequence in task memory and reusable recovery rules in global memory.
At test time, it retrieves this memory, locates targets in RGB-D observations, and uses fixed analytic primitives for positioning and the frozen policy for contact. Failed contacts can be retried within the episode without resets.
RMBench-Compose compares $\pi_{0.5}$~\citep{physicalintelligence2025pi05}, HarnessVLA, Orchestra with $\pi_{0.5}$, the complete Orchestra system, and its Qwen3.7-Plus-agent variant.
\emph{Orchestra with $\pi_{0.5}$} uses the base action policy instead of the spatially and temporally conditioned policy in Section~\ref{sec:orchestra-policy}.
The Qwen-agent variant replaces only the task-orchestration agent. 
The supplementary material lists the sources of the baseline results and describes the model configurations.

\textbf{Implementation details.}
The temporally conditioned action policy uses a visual history window of \(K=6\) frames with a sampling stride of \(\Delta=3\).
The progress evaluator uses \(B=11\) discrete within-subtask progress categories.
The appendix describes how we construct the progress targets and lists the training hyperparameters for both the action policy and the progress evaluator.

% \textbf{Metrics.}
% We report task success rates on RMBench~\citep{chen2026rmbench} and Atomic SR, Subtask Success Rate (SSR), and Average Success Length (ASL) on RMBench-Compose.
% Atomic SR measures isolated-task success; SSR measures success among attempted subtasks, excluding those unreached after an earlier failure.
% ASL is the mean number of consecutive successful subtasks in the requested order, bounded by the task length $n$.

% \input{tables/rmbench_results}

\subsection{Evaluation on RMBench}
\label{sec:results-rmbench}

Orchestra achieved the highest mean success rate across the nine tasks ($68.22\%$ versus $42.00\%$ for Mem-0, Table~\ref{tab:orchestra-rmbench}).
Nearly three quarters of the lead over Mem-0 came from two tasks that require the robot to remember a target's identity or keep count across repeated actions~\citep{chen2026rmbench}. On Observe and Pick Up, Orchestra picked up the object matching a hidden reference in $81\%$ of trials, compared with $4\%$ for Mem-0 and $10\%$ for HarnessVLA. On Press Button, it pressed the buttons the prescribed number of times in $97\%$ of trials, compared with $0\%$ and $44\%$, respectively.
Across the other seven tasks, Orchestra still led Mem-0 by $8.9$ percentage points on average. Mem-0 remained stronger on Put Back Block ($90\%$ versus $81\%$), and Swap T remained difficult for every method (at most $20\%$ success, including $10\%$ for Orchestra).

\subsection{Compositional Execution on RMBench-Compose}
\label{sec:results-compose}

\textbf{Execution with a fixed action policy.}
The base $\pi_{0.5}$ policy succeeded on $85.0\%$ of isolated trials but averaged less than one completed subtask ($0.93$ ASL) when given the full instruction (Fig.~\ref{fig:results-compose}).
HarnessVLA and Orchestra retained the same Atomic SR with this policy, yet reached $2.99$ and $3.60$ ASL, respectively. SSR likewise rose from $47.4\%$ to $72.1\%$ and $76.0\%$.
By selecting the active subtask and deciding when to advance, Orchestra let the same policy complete nearly four times as many consecutive subtasks as it did when given the full instruction directly.

\textbf{Execution with spatial and temporal policy conditioning.}
With orchestration fixed, spatial prompts and recent frames improved the policy's success on both isolated skills ($85.0\%$ to $95.2\%$ Atomic SR) and attempted subtasks in a sequence ($76.0\%$ to $86.7\%$ SSR). With more individual steps succeeding, the conditioned policy completed more than twice as many consecutive subtasks ($7.35$ versus $3.60$ ASL).
The full system exceeded HarnessVLA in ASL in all five environments ($2.46\times$ on average), reaching $13.06$ of $18$ steps on Place Block on Mat versus $4.35$ for HarnessVLA.
Orchestra came close to completing the full Press Button sequence ($5.79$ of $6$ steps) but averaged only $2.11$ of $10$ on Pick Up Object. The robot thus progressed further through the 18-step placement task than through the shorter pickup task, suggesting that the type of manipulation also matters for sustained execution.

\textbf{Alternative orchestration agent.}
Replacing the default agent with Qwen yielded $85.7\%$ SSR and $7.19$ ASL, compared with $86.7\%$ and $7.35$ for the default agent. Atomic SR remained $95.2\%$. Qwen led narrowly on Cover and Uncover Block ($5.90$ versus $5.82$ ASL). Across these environments, changing the policy's spatial and temporal inputs had a larger effect on sustained execution than changing which of these two agents supplied the subtasks.
% The supplementary material reports Atomic SR and SSR for each environment.

\begin{figure}[t]
    \centering
    \includegraphics[width=\columnwidth]{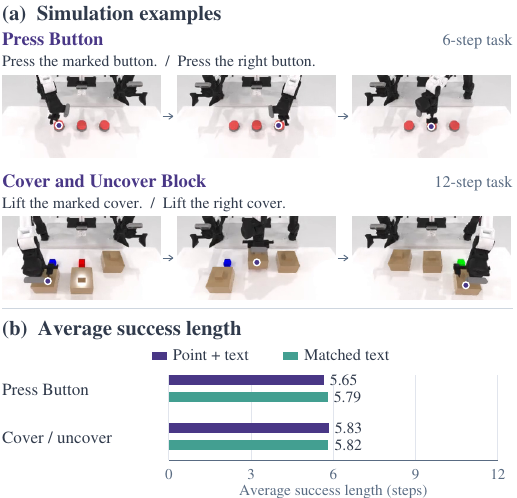}
    \caption{\textbf{Point prompts and matched text instructions on RMBench-Compose.}
    \textbf{(a)} Dots mark action targets in successive snapshots. Directions are robot-relative.
    \textbf{(b)} Average success length (ASL) over 100 episodes per task and input condition.}
    \label{fig:visual-instruction-asl}
\end{figure}

\textbf{Following multimodal instructions.}
We keep the requested actions fixed and identify the same target either with a simulator-generated point in the initial image or with a text description (Fig.~\ref{fig:visual-instruction-asl}). ASL with point-plus-text and text alone was $5.65$ and $5.79$ on Press Button, and $5.83$ and $5.82$ on Cover and Uncover Block.
Orchestra thus followed the same action sequence to a similar depth whether the instruction identified the target by a point or by its location in words. This held both near the six-step maximum on Press Button and roughly halfway through the 12-step Cover and Uncover Block sequence.

\subsection{Ablation Studies}
\label{sec:results-ablation}

\begin{figure*}[t]
\centering
\includegraphics[width=\textwidth]{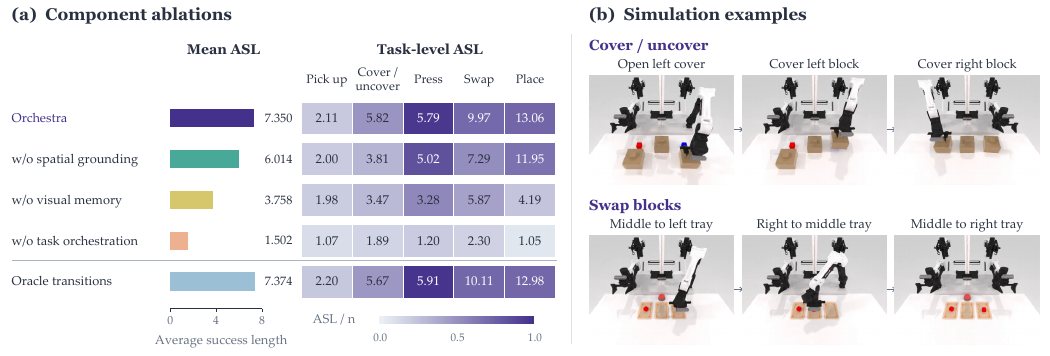}
% \vsapce{-3mm}
\caption{\textbf{Component ablations and simulation examples on RMBench-Compose.} \textbf{(a)} Mean and per-environment ASL for component removals and state-based oracle transitions. Cells report ASL in subtasks, and shading shows the fraction of task length completed (ASL/$n$). \textbf{(b)} Selected subtask frames illustrating Cover and Uncover Block and Swap Block.}
\vspace{-3mm}
\label{fig:results-ablation}
\end{figure*}

\textbf{Component contributions.}
Removing task orchestration produced the largest drop in mean ASL ($7.350$ to $1.502$, Fig.~\ref{fig:results-ablation}a). Even with training demonstrations matched to the full instructions, asking the policy to act on the entire instruction was a poor substitute for giving it one active subtask at a time.
Removing recent frames from the action policy alone reduced mean ASL to $3.758$. Place Block on Mat was especially sensitive to the policy's visual history. From a baseline of $13.06$ steps, ASL fell to $4.19$ without recent frames but remained at $11.95$ without spatial grounding. The evaluator retained its history in the first ablation, so the larger drop isolates the value of recent observations for choosing the next movement.
Removing spatial grounding reduced mean ASL to $6.014$. Target locations mattered most on Swap Block ($9.97$ to $7.29$) and Cover and Uncover Block ($5.82$ to $3.81$), where successive instructions refer to different blocks, trays, or covers in the same scene.

\textbf{Learned versus state-based transitions.}
Orchestra sustained nearly the same sequence length with learned transitions as with exact simulator feedback. Replacing the learned evaluator with the simulator's completion indicator changed mean ASL from $7.350$ to $7.374$, with a gap of at most $0.15$ in any environment.
Even with exact completion information, Pick Up Object reached only $2.20$ of $10$ steps. The oracle can advance after a successful pickup, but the policy must still execute the grasp within its action budget. Better switch timing alone therefore leaves most of the execution shortfall on this task unresolved.

\begin{figure*}[ht]
\centering
\includegraphics[width=\textwidth]{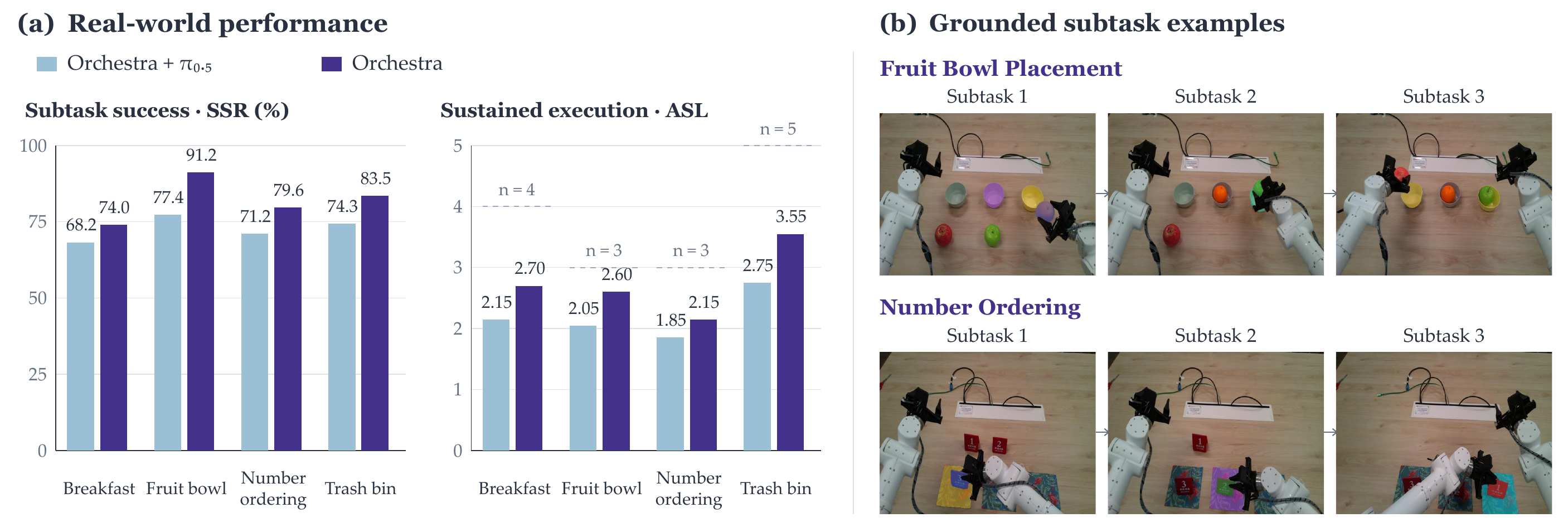}
% \vsapce{-3mm}
\caption{\textbf{Real-world performance and grounded subtask examples.} \textbf{(a)} SSR and ASL on four tasks over $20$ episodes per task and configuration. Dashed lines mark task lengths ($n=4,3,3,5$). 
% Both configurations share the Orchestra harness; the comparison tests the complete system against its base-$\pi_{0.5}$-policy configuration. 
\textbf{(b)} Selected subtask examples for Fruit Bowl Placement and Number Ordering, with corresponding object masks overlaid on RGB frames. 
% Each row follows the subtask order and comprises separately recorded examples.
}
\label{fig:results-realworld}
% \vspace{-3mm}
\end{figure*}

\textbf{Visual history improves subtask switch timing.}
Across 500 cases per method, six frames of history raised the evaluator's on-time switch rate from $77.0\%$ to $96.6\%$ against the simulator oracle (Table~\ref{tab:progresser-eval}). History reduced both early switches ($12.8\%$ to $2.6\%$) and false switches on timeout ($10.2\%$ to $0.8\%$).
The binary classifier labels only the final progress bin ($10/11<\rho_i(u)\leq 1$) as done. It switched early least often ($1.2\%$), yet falsely switched on timeout in $43.4\%$ of cases. Together with the policy ablation, these results suggest two roles for recent observations in Orchestra. They help the policy choose its next movement and the evaluator decide when to advance to the next subtask.

\begin{table}[ht]
\centering
\caption{\textbf{Subtask switch timing.} With six frames of visual history, our progress evaluator switches on time most often among the learned methods. Rates are percentages over 500 cases per method.}
\label{tab:progresser-eval}
\small
\setlength{\tabcolsep}{2pt}
\begin{tabularx}{\linewidth}{@{}>{\raggedright\arraybackslash}Xrrr@{}}
\toprule
Transition signal & Early $\downarrow$ & On-time $\uparrow$ & \shortstack[r]{False switch\\on timeout $\downarrow$} \\
\midrule
Oracle transitions & 0.0 & 100.0 & 0.0 \\
\midrule
Task-done classifier & \textbf{1.2} & 55.4 & 43.4 \\
Evaluator (current frame) & 12.8 & 77.0 & 10.2 \\
Evaluator (6-frame history) & 2.6 & \textbf{96.6} & \textbf{0.8} \\
\bottomrule
\end{tabularx}
\end{table}

\subsection{Real-World Evaluation}
\label{sec:results-realworld}

We evaluated Breakfast Preparation, Fruit Bowl Placement, Number Ordering, and Trash Bin Placement, with instruction lengths of $4$, $3$, $3$, and $5$ subtasks, respectively.
We ran $20$ evaluation rollouts per task for each configuration, totaling $80$ rollouts per configuration.
Both configurations used the same set of randomly selected task instructions.
Execution used predicted and tracked grounding on the robot platform described in the supplementary material.

With the Orchestra harness fixed, the spatially and temporally conditioned policy improved SSR and ASL on all four tasks (Fig.~\ref{fig:results-realworld}a). Mean SSR rose from $72.77\%$ to $82.09\%$, with the largest gain on Fruit Bowl Placement ($77.36\%$ to $91.22\%$). Using targets predicted and tracked on the robot, the conditioned policy completed $25\%$ more consecutive subtasks on average ($2.75$ versus $2.20$ ASL). The largest gain in sequence length came on five-step Trash Bin Placement, where the robot completed nearly one additional placement per instruction ($3.55$ versus $2.75$).

The supplementary material presents additional qualitative real-world results and analyzes failure cases.

\section{Discussion and Future Work}
\label{sec:conclusion}

\noindent\textbf{Composing atomic skills through subtask orchestration.}
% Orchestra lets a fixed action policy execute longer sequences by deciding where to act and when to advance. This still requires the plan to cover every requested object and action.
Orchestra uses the active subtask to decide where to act and when to advance. On RMBench-Compose, the same $\pi_{0.5}$ policy averaged $0.93$ consecutive subtasks when given the full instruction and $3.60$ when Orchestra managed the sequence. This gain comes from coordinating skills the policy has already learned. The plan must still cover every requested object and action.

\noindent\textbf{Future work --- revising the plan as the scene changes.}
After each subtask attempt, Orchestra could check the latest observation against the original instruction to identify unmet goals. It could then update target locations and revise the remaining steps to address them in the requested order.

\section*{LLM Usage Statement}
We used generative AI tools to polish author-written text, assist with parts of the experimental and analysis code, and help produce figures from our results. All AI-assisted work was reviewed by the authors, who take full responsibility for the final text, claims, code, and figures.

{\small
\bibliographystyle{ieeenat_fullname}
\bibliography{main}
}

\onecolumn

\clearpage
\appendix

\section*{Appendix Overview}

The appendix is organized as follows:
\begin{itemize}
    \item \textbf{Appendix~\ref{app:agent-framework}: Agent Framework and Execution Details.}
    Describes task decomposition, the agent prompt and output schema, shared context, memory management, and progress-driven execution.

    \item \textbf{Appendix~\ref{app:orchestra-B}: Policy and Evaluator Implementation.}
    Details the action policy, spatial conditioning, short-horizon visual memory, and progress evaluator, including progress-target generation and subtask transitions.

    \item \textbf{Appendix~\ref{app:orchestra-A}: Data Collection and Training.}
    Summarizes real-world and simulation demonstration collection, annotation procedures, and training configurations.

    \item \textbf{Appendix~\ref{app:evaluation-protocols}: Evaluation Protocols.}
    Specifies composed instructions, execution budgets, evaluation metrics, baseline sources, and the physical robot platform.

    \item \textbf{Appendix~\ref{app:additional-results}: Additional Results and Analysis.}
    Presents per-environment compositional results, ablation implementation details, qualitative execution examples, and failure analysis.

\end{itemize}

\section{Agent Framework and Execution Details}
\label{app:agent-framework}
\label{app:orchestra-D}

\subsection{Task interpretation and registered capabilities}

Orchestra receives the complete user instruction and the current scene, together with the registered action-policy capabilities. The instruction may be language-only or include a reference image, points, or bounding boxes. The agent converts these inputs into an ordered sequence of executable subtasks and identifies the entities relevant to each subtask. The complete instruction is available at the beginning of an episode.
% ; the environment does not supply the next instruction incrementally to Orchestra.

Unless otherwise specified, the task-orchestration agent is DeepSeek-V4.1-Flash. The Qwen3.7-Plus variant replaces this orchestration agent, as evaluated in Section~\ref{sec:results-compose}. The framework exposes action generation and progress detection as separate capabilities. The experiments instantiate them with the policy and evaluator described below; the shared-context interface permits other input configurations without changing the overall orchestration structure.

\subsection{Agent prompt and output schema}
\label{app:agent-prompt}

\begin{figure}[t]
    \centering
    \includegraphics[width=\linewidth]{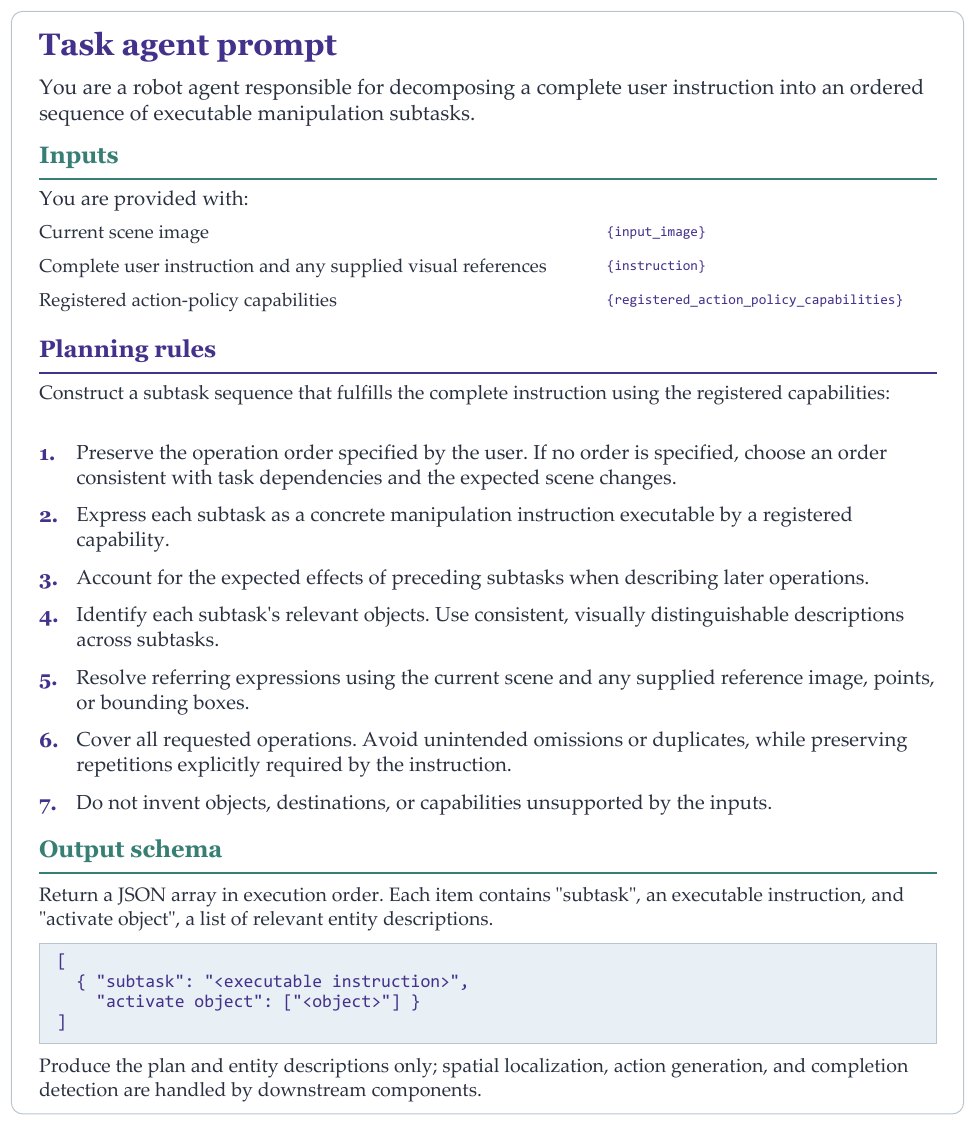}
    \caption{
        \textbf{Task-orchestration agent prompt and output schema.}
        The template specifies the scene, instruction, and capability inputs; seven rules for subtask decomposition and entity identification; and the required JSON output format. Each output entry pairs an executable manipulation instruction with descriptions of its relevant entities. Spatial localization, action generation, and completion detection are delegated to downstream components.
    }
    \label{fig:agent-prompt}
\end{figure}

Figure~\ref{fig:agent-prompt} presents the prompt template used by the task-orchestration agent to decompose a complete user instruction into an ordered sequence of manipulation subtasks. The agent receives the current scene image, the user instruction with any supplied visual references, and the registered action-policy capabilities. The prompt directs the agent to preserve explicit operation order, account for task dependencies and expected scene changes, and identify relevant entities using consistent, visually distinguishable descriptions. It also requires coverage of all requested operations, including explicitly specified repetitions, without introducing unsupported objects, destinations, or capabilities.

The output is a JSON array in execution order. Each entry contains a \texttt{subtask} field specifying a manipulation instruction and an \texttt{activate object} field listing the relevant entity descriptions. This interface supplies the subtask sequence and textual entity descriptions to downstream components, which perform spatial localization, action generation, and completion detection.

\subsection{Shared context and spatial grounding}

The shared context contains the active instruction, current observations, recent visual history, task records, and grounded entities. At a subtask boundary, MolmoPoint-8B localizes the relevant entities in the head-camera image and SAM3 initializes their masks. At subsequent action-chunk replanning calls, the segmentation session tracks those entities. A subtask transition resets the grounding session for the next instruction.

Both the action policy and the progress evaluator use observations from the shared context. They maintain independent model weights and temporal buffers. Spatial representation choices are detailed in Appendix~\ref{app:orchestra-B}.

\subsection{What the agent remembers}

\begin{center}
\small
\begin{tabularx}{\linewidth}{@{}XXX@{}}
\toprule
Memory content & Role & Boundary \\
\midrule
Completed-subtask names and start/end frames & Explicit execution records & Retained across subtasks; cleared for a new episode \\
Current plan and active index & Identify the current and remaining goals & Advance at terminal-bin prediction \\
Timestamped multi-view RGB history & Supply recent local visual observations & Bounded; the next subtask's local window is reinitialized \\
Grounded entity data & Associate the active goal with current image regions & Recomputed for the next subtask \\
Policy/evaluator observation buffers & Construct model-specific temporal input & Separate buffers reset by the agent \\
\bottomrule
\end{tabularx}
\end{center}

The agent explicitly stores the name and the start and end frames of each completed subtask. These records are retained across subtask transitions and cleared for a new episode. Together, they provide an explicit episode-level execution trace of completed goals and their temporal boundaries.

The recent observation history supplies temporal visual inputs to the policy and evaluator. The current proprioceptive state is passed with the current observation. Grounding sessions and model-side buffers follow the active subtask's lifecycle.

\subsection{Progress-driven runtime algorithm}

As shown in Algorithm~\ref{alg:orchestra-runtime}, Orchestra
executes the subtasks in order and uses predicted progress
to determine when to advance to the next subtask. At each replanning call, the policy predicts an action chunk while the evaluator estimates progress from the same observation context. If the predicted category is terminal, the agent discards the chunk and advances the plan. Otherwise, it executes the first $H_e=30$ actions of the $H_p=50$ predicted actions, or fewer if the current subtask's remaining budget is smaller.

\begin{algorithm}[t]
\caption{Progress-driven execution in Orchestra}
\label{alg:orchestra-runtime}
\small
\begin{algorithmic}[1]
\Require Instruction $\ell$, observation stream,
         policy $\pi_\theta$, evaluator $e_\phi$
\Ensure Termination status and completed-subtask records $\mathcal D$

\State Decompose $\ell$ into an ordered plan
       $\mathcal G=(g_1,\ldots,g_N)$
\State $\mathcal D \gets \varnothing$; $i \gets 1$
\State Clear visual history, model buffers, and grounding
\State Record the start frame of $g_i$

\State Set $r$ to the action budget for $g_i$
\While{$r>0$}
    \State Obtain current RGB images and proprioceptive state
    \State Update visual history and model buffers
    \State Initialize or update grounding $F$ for $g_i$
    \State Construct the shared context $C$ for $g_i$
    \State Run in parallel:
        $A \gets \pi_\theta(\Pi_\pi(C))$,
        $q \gets e_\phi(\Pi_e(C))$

    \If{$\operatorname*{arg\,max}_{b\in\{0,\ldots,10\}} q_b = 10$}
        \State Append $(g_i,\text{start frame},\text{current frame})$
               to $\mathcal D$
        \State Discard $A$ and any pending actions for $g_i$
        \If{$i=N$}
            \State \Return
                $(\texttt{PLAN\_TERMINATED},\mathcal D)$
        \EndIf
        \State $i \gets i+1$ and reset $r$ to the budget for the new subtask
        \State Clear visual history, model buffers, and grounding
        \State Record the current frame as the start frame of $g_i$
        \State Seed the new visual history with the current images
        \State \textbf{continue} with a new inference call for $g_i$
    \Else
        \State Execute the first $k=\min(H_e,r)$ actions of $A$
        \State $r \gets r-k$
        \State Update visual history and model buffers
               as observations arrive
    \EndIf
\EndWhile

\State \Return $(\texttt{BUDGET\_EXHAUSTED},\mathcal D)$
\end{algorithmic}
\end{algorithm}

Intermediate observations update the RGB histories without triggering grounding or model inference. Episode and observation indices ensure that each image is appended only once. Proprioception is supplied as the current state and is not stored in the visual history. At each subtask transition, the agent resets the local histories and grounding, while retaining completed-subtask records for bookkeeping.

The evaluator assesses the observed subtask stage before the newly predicted action chunk is executed. A terminal prediction therefore triggers a subtask transition based on the current observations. After the final subtask, the agent terminates the plan.

\section{Policy and Evaluator Implementation}
\label{app:orchestra-B}

\subsection{Action-policy architecture}

The action policy is initialized from the $\pi_{0.5}$ base checkpoint and combines the Gemma-2B language backbone within PaliGemma with an action expert using OpenPI's \texttt{gemma\_300m} configuration. It predicts action chunks from the active subtask, spatial conditions, recent visual observations, and current proprioception. Temporal attention reuses spatial-attention parameters, and the temporal positional encodings are fixed buffers. Thus, the memory pathway adds no new learned memory parameters. Training and execution settings are listed in Appendix~\ref{app:orchestra-A}.

\subsection{Entity selection and spatial prompts}

During training, per-frame active-entity annotations select the objects relevant to the current language subtask. End-effector entities are excluded. 
When spatial conditioning is retained, the transform samples one of three representations with weights 0.34, 0.33, and 0.33:

\begin{center}
\small
\begin{tabularx}{\linewidth}{@{}XXX@{}}
\toprule
Representation & Image processing & Text processing \\
\midrule
Contour & Overlay object contours on the current head-camera image & Append entity bounding boxes \\
Bounding box & Leave RGB unchanged & Append named bounding-box coordinates \\
Point & Leave RGB unchanged & Append named point coordinates \\
\bottomrule
\end{tabularx}
\end{center}

Coordinates are quantized to 256 bins. The contour configuration uses green boundaries with blending coefficient 0.4. The policy drops all spatial conditions with probability 0.2 to encourage robustness to missing or inaccurate grounding information. Policy training also uses image dropout with probability 0.1 to encourage greater reliance on grounding information rather than solely on image observations. In the image-drop branch, wrist views are retained with probability 0.9. These augmentations vary the availability of visual and spatial inputs during training.

\paragraph{Inputs at inference.}
At inference, the action policy encodes a six-frame window for each camera. The current head-camera frame shows the tracked contours of the active subtask's entities. Earlier head-camera frames and all wrist-camera frames retain their original RGB values. The policy also receives the entity names and quantized bounding boxes alongside the active instruction. The progress evaluator uses the same current-frame contour overlay in its own visual window. Training dropouts are disabled.

\subsection{Temporal window and encoder}

\begin{figure}[ht]
    \centering
    \includegraphics[width=0.95\linewidth]{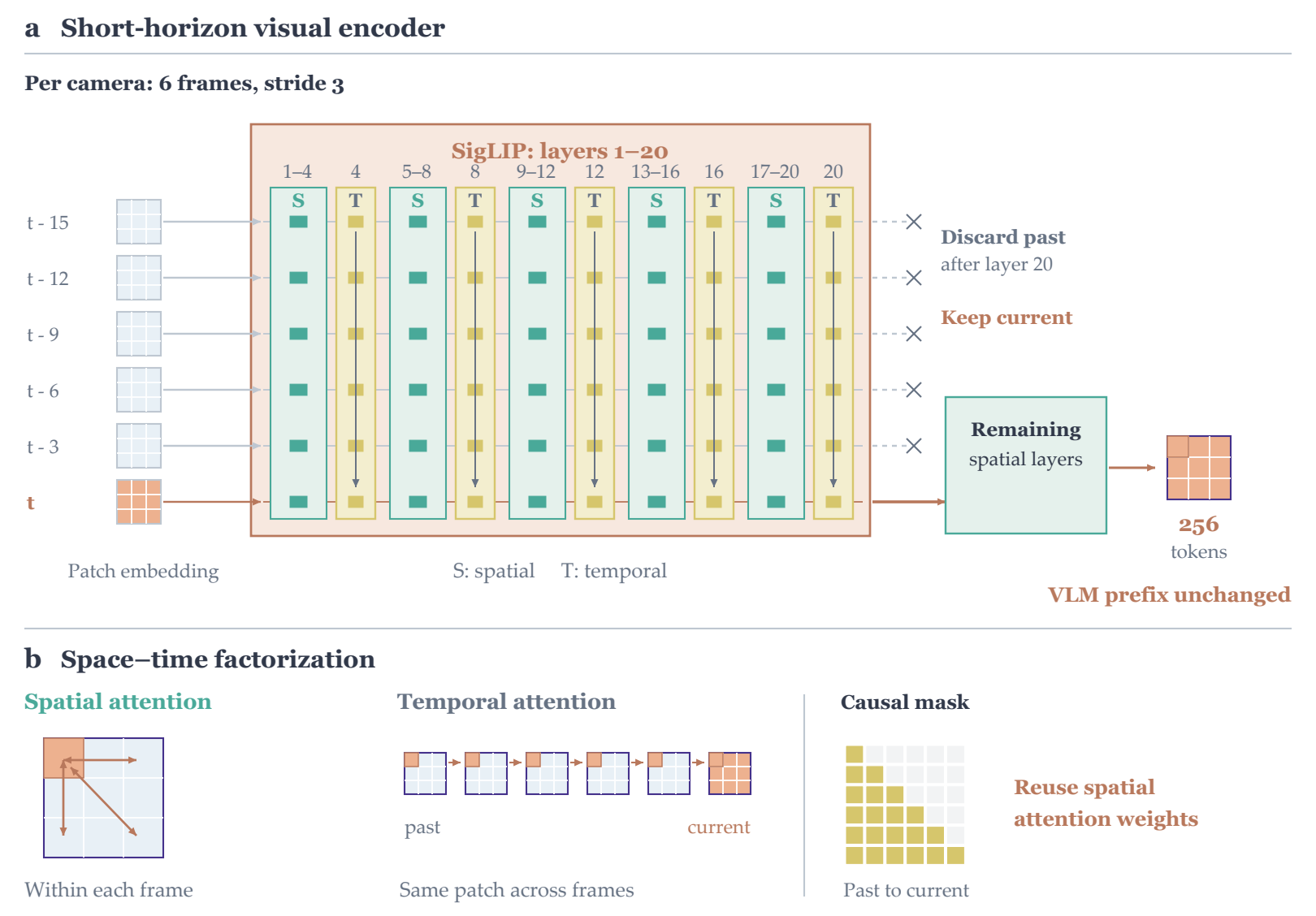}
    \caption{\textbf{Short-horizon visual memory inside SigLIP.}
    \textbf{(a)} Each camera supplies a six-frame window with stride three, spanning observation indices $t-15$ through $t$. At inference, the current head-camera frame includes the tracked entity contours. Patch embedding and spatial positional encoding are applied independently to each frame. Four spatial layers (S) alternate with a causal temporal-attention operation (T), giving temporal mixing after layers 4, 8, 12, 16, and 20. After layer 20, historical tokens are discarded and only current-frame features pass through the remaining spatial layers. The VLM receives 256 visual tokens per camera, matching the single-frame baseline.
    \textbf{(b)} Spatial attention is bidirectional within a frame; temporal attention operates across frames at the same patch index, with causal and valid-history masks. The temporal pathway reuses the corresponding spatial-attention weights and adds fixed sinusoidal age encodings, with zero encoding for the current frame. This adaptation introduces no new learned memory parameters. Patch grids are schematic.}
    \label{fig:short-horizon-memory}
\end{figure}

Figure~\ref{fig:short-horizon-memory} details our instantiation of the space--time factorization used in the policy's visual encoder.

Each camera contributes six observations separated by three observation indices, including the current image. We report the window configuration in observation indices.

Frames are embedded spatially before temporal mixing. Temporal attention is inserted after layers 4, 8, 12, 16, and 20 of the visual encoder, using one-based layer numbering. For each patch position, causal attention operates across frame indices, masks unavailable history, and reuses the corresponding layer's attention parameters. Fixed sinusoidal temporal encodings express frame age. Following layer 20, past-frame tokens are discarded, leaving the current-frame patches for subsequent visual layers.

At the documented resolution and patch size, the language model receives 256 visual tokens per camera, as in the single-frame baseline. History increases the visual encoder's work, while leaving this visual prefix count unchanged. Stride changes the sampled temporal span without changing the number of frames processed.

\begin{figure}[ht]
    \centering
    \includegraphics[width=0.8\linewidth]{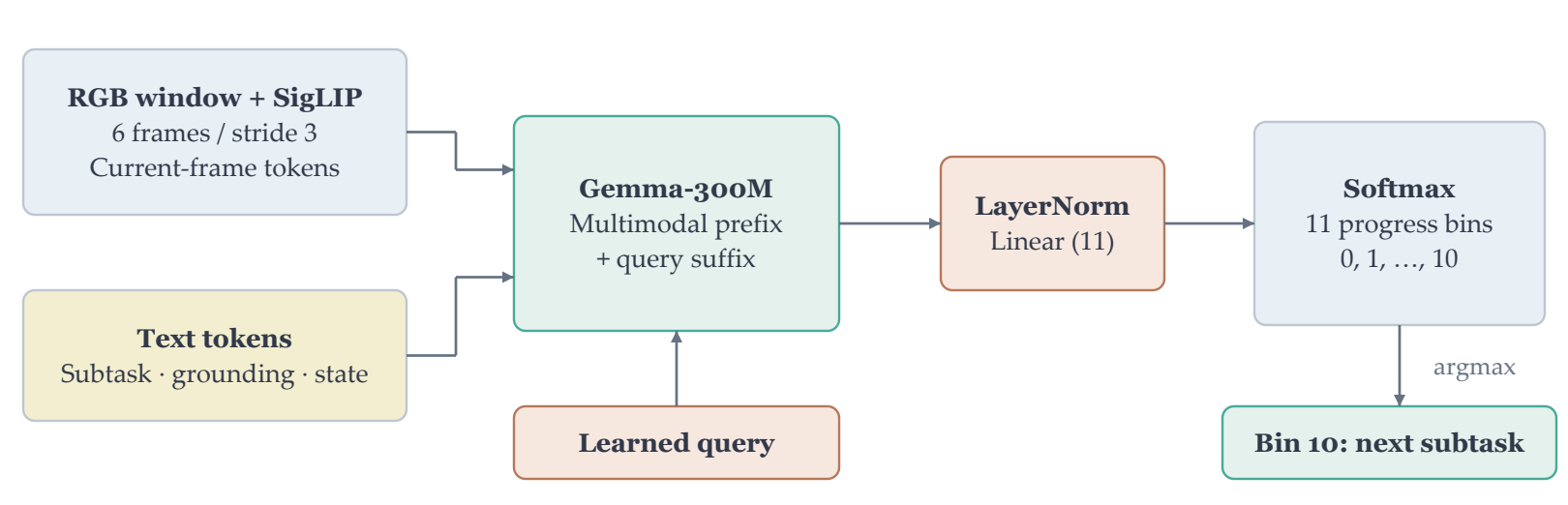}
    \caption{\textbf{Architecture of the progress evaluator.} A learned query aggregates the multimodal visual and language prefix, and a normalized linear head predicts one of 11 progress categories.}
    \label{fig:progress-evaluator}
\end{figure}

\subsection{Progress-evaluator configuration}

The progress evaluator pairs a visual encoder with a language backbone built with the \texttt{gemma\_300m} configuration from OpenPI~\citep{physicalintelligence2025pi05}. A learned query of width 1024 aggregates information from the multimodal prefix. Layer normalization and a linear projection produce the 11 progress logits. The evaluator uses the same six-frame, stride-three observation window as the policy and has independent weights.

The deployed subtask transition uses only Eq.~\eqref{eq:orchestra-8}, while Eq.~\eqref{eq:orchestra-7} specifies the progress classification objective.

\subsection{Automatically generated progress targets and transitions}
\label{app:orchestra-C}

Progress targets are generated automatically from the step index within each demonstration subtask; no manual progress-category labels are required. For a subtask segment of \(T_i=e_i-a_i\ge1\) frames, Eq.~\eqref{eq:orchestra-6} 
reduces to

\begin{equation}
\label{eq:orchestra-9}
\rho_i(u)=1-\frac{\max(e_i-1-u,0)}{T_i}.
\end{equation}

The first frame has \(\rho_i=1/T_i\), and the last has \(\rho_i=1\). We divide \((0,1]\) into 11 equal-width intervals. Frame \(u\) receives label \(k\) when \(k/11<\rho_i(u)\leq(k+1)/11\), for \(k=0,\ldots,10\). The task-done classifier labels a frame complete exactly when \(k=10\). For example, in a 110-frame segment, counting from one, frames 101 through 110 receive the terminal label. During deployment, the agent uses the argmax class directly. A prediction of bin 10 triggers the transition.

For visualization, we assign each class \(j\) the display value \(j/10\), mapping the first and last classes to 0 and 1. The expected normalized class index under \(q_t\) is

\begin{equation}
\label{eq:orchestra-10}
\begin{gathered}
s_t^{\mathrm{vis}}=\sum_{j=0}^{10}\frac{j}{10}q_{t,j}.
\end{gathered}
\end{equation}

\section{Data Collection and Training}
\label{app:orchestra-A}

\subsection{Real-world training data}

We collected demonstrations for Breakfast Preparation, Fruit Bowl Placement, Number Ordering, and Trash Bin Placement at 20~Hz. Demonstrations from all four tasks were combined for joint training. Collection was performed at the subtask level; Table~\ref{tab:app-real-data} reports the amounts using a complete-task accounting unit, where one unit corresponds to one completion of the task. 

\begin{table}[ht]
\centering
\small
\caption{Real-world data collection and evaluation. Collection amounts use complete-task accounting units assembled from subtask-level collection. Evaluation counts are per configuration.}
\label{tab:app-real-data}
\begin{tabularx}{\linewidth}{@{}Xrrr@{}}
\toprule
Task & Collection units & Eval. episodes & Eval. length \\
\midrule
Breakfast Preparation & 120 & 20 & 4 \\
Fruit Bowl Placement & 120 & 20 & 3 \\
Number Ordering & 144 & 20 & 3 \\
Trash Bin Placement & 120 & 20 & 5 \\
\bottomrule
\end{tabularx}
\end{table}

For Number Ordering, collection covered all six permutations of the three numbers, with 24 demonstrations per order under the same accounting convention. The real-world annotation pipeline uses Qwen3.7-Plus for subtask decomposition, MolmoPoint for object pointing, and SAM3 for segmentation. Progress categories are generated automatically from the step position within each demonstration subtask, as described in Appendix~\ref{app:orchestra-C}.

\subsection{RMBench-Compose training data}

For each environment, let $m$ denote the number of task steps required to complete one operation for every object in the environment. We use this quantity to set the collection scale: we collected $50m$ atomic demonstrations and set the composed evaluation length to $2m$. Training uses atomic instructions and subtask demonstrations, whereas evaluation supplies a complete instruction. Table~\ref{tab:app-sim-protocol} reports the environment-specific amounts and evaluation budgets.

\subsection{Training configuration}

Unless otherwise specified, the action policy and progress evaluator use the same training hyperparameters, including 50,000 training steps and a global batch size of 20 across 2 Nvidia A6000 Pro GPUs. They have independent weights and distinct training objectives: flow matching for action prediction and cross-entropy for progress classification. Architecture-specific settings and action horizons apply only to the corresponding model.

\begin{center}
\small
\begin{tabularx}{\linewidth}{@{}XX@{}}
\toprule
Setting & Configuration \\
\midrule
Policy initialization & \(\pi_{0.5}\) base checkpoint \\
Policy language backbone & Gemma-2B within PaliGemma \\
Policy action expert & OpenPI \texttt{gemma\_300m} \\
Maximum text sequence length & 300 \\
Policy prediction / execution horizons & 50 / 30 steps \\
Precision & bfloat16 \\
Global batch size (across both GPUs) & 20 \\
Training steps & 50,000 \\
Optimizer & AdamW, betas 0.9 and 0.95, epsilon \(10^{-8}\) \\
Peak / final learning rate & \(2.5\times10^{-5}\) / \(2.5\times10^{-6}\) \\
Warmup / schedule decay steps & 1,000 / 30,000 \\
% Training hardware & Two A6000 Pro GPUs \\
Weight decay / gradient clipping & \(10^{-10}\) / global norm 1.0 \\
\bottomrule
\end{tabularx}
\end{center}

\section{Evaluation Protocols}
\label{app:evaluation-protocols}

\subsection{Composed instructions and execution budgets}

For each RMBench-Compose environment and evaluated configuration, we ran 100 episodes. Each episode began with a randomly selected complete instruction of the environment's specified maximum evaluation length. Orchestra received the entire instruction and decomposed it into subtasks before sequential execution. Scene states persisted across subtask boundaries. These experiments use one specified evaluation length per environment.

\begin{table}[ht]
\centering
\small
\caption{RMBench-Compose collection and evaluation settings. $m$ is the minimum task length used for collection scaling; $n=2m$ defines a deliberately long evaluation sequence designed to stress-test the policy's limits in sustained execution. Budgets are maximum executed action steps per subtask, not inference-call counts. Each environment is evaluated for 100 episodes per configuration.}
\label{tab:app-sim-protocol}
\begin{tabularx}{\linewidth}{@{}Xrrrr@{}}
\toprule
Environment & $m$ & Demos ($50m$) & $n$ & Budget \\
\midrule
Pick Up Object & 5 & 250 & 10 & 250 \\
Cover and Uncover Block & 6 & 300 & 12 & 300 \\
Press Button & 3 & 150 & 6 & 150 \\
Swap Block & 7 & 350 & 14 & 300 \\
Place Block on Mat & 9 & 450 & 18 & 250 \\
\bottomrule
\end{tabularx}
\end{table}

In complete Orchestra, the evaluator's terminal-category prediction triggers subtask advancement. Environment-state completion replaces this learned signal only in the oracle-transition ablation in Section~\ref{sec:results-ablation}. Environment success predicates are used for external scoring and should not be interpreted as an additional switching condition in the complete system. 

\subsection{Metrics and real-world evaluation}

% Atomic Success Rate (Atomic SR) measures isolated atomic-task execution. Subtask Success Rate (SSR) is the fraction of attempted subtasks completed successfully, excluding subtasks not reached after an earlier failure. Average Success Length (ASL) measures the mean number of consecutive subtasks completed in the requested order and is bounded by the instruction length. Cross-environment averages are unweighted arithmetic means. Any subtask failure terminates the corresponding evaluation rollout.

For each real-world task, we randomly selected 20 instructions and evaluated both policy configurations on the same instruction set, with one evaluation per selected instruction and configuration. This gives 80 evaluations per configuration across the four tasks. Data collection and robot execution both operate at 20~Hz. Any subtask failure terminates the corresponding evaluation rollout.

\subsection{Baseline comparability}
\label{app:baseline-sources}

% For the RMBench simulation comparison in Table~\ref{tab:orchestra-rmbench}, we use the officially reported RMBench results for nine baselines. These comprise DP, ACT, $\pi_{0.5}$, X-VLA, QwenOFT, MemER, Mem-0, MemoryVLA (OpenVLA), and MemoryVLA (QwenOFT). For HarnessVLA (Codex), we reproduce the baseline on RMBench using the HarnessVLA source code. The table therefore combines official benchmark results with our evaluations of HarnessVLA (Codex) and Orchestra.

For the RMBench simulation comparison in Table~\ref{tab:orchestra-rmbench}, we use the officially reported RMBench results for nine baselines. These comprise DP~\citep{chi2023diffusionpolicy}, ACT~\citep{zhao2023act}, $\pi_{0.5}$~\citep{physicalintelligence2025pi05}, X-VLA~\citep{zheng2025xvla}, QwenOFT~\citep{starvla2026}, MemER~\citep{sridhar2025memer}, Mem-0~\citep{chen2026rmbench}, MemoryVLA (OpenVLA)~\citep{shi2025memoryvla}, and MemoryVLA (QwenOFT)~\citep{shi2025memoryvla,yang2026eventvla}. For HarnessVLA (Codex)~\citep{zhang2026harnessvla}, we reproduce the baseline on RMBench using the HarnessVLA source code. HarnessVLA uses Codex as its default agent. The table therefore combines official benchmark results with our evaluations of HarnessVLA (Codex) and Orchestra.

For Orchestra on RMBench, the task-orchestration agent receives the complete episode instruction, the associated task description, and the initial scene image. It generates an ordered sequence of subtask instructions together with descriptions of the relevant entities. For example, for the generated subtask ``open the left cover,'' the agent identifies ``the left cover'' as the relevant entity.

For RMBench, we collect a demonstration dataset matched in size to the original RMBench training data and organize it into subtask-level training examples. We use these same demonstrations to train both the $\pi_{0.5}$ action policy in HarnessVLA and the Orchestra action policy. On RMBench-Compose, both policies likewise use the same atomic demonstrations collected for that benchmark. The results in Table~\ref{tab:orchestra-rmbench} compare the full systems, including differences in spatial inputs, policy architecture, and execution management.

\subsection{Physical robot platform}
\label{app:robot-platform}

\begin{wrapfigure}{r}{0.45\linewidth}
    \centering
        \vspace{-3mm}
    \includegraphics[width=\linewidth]{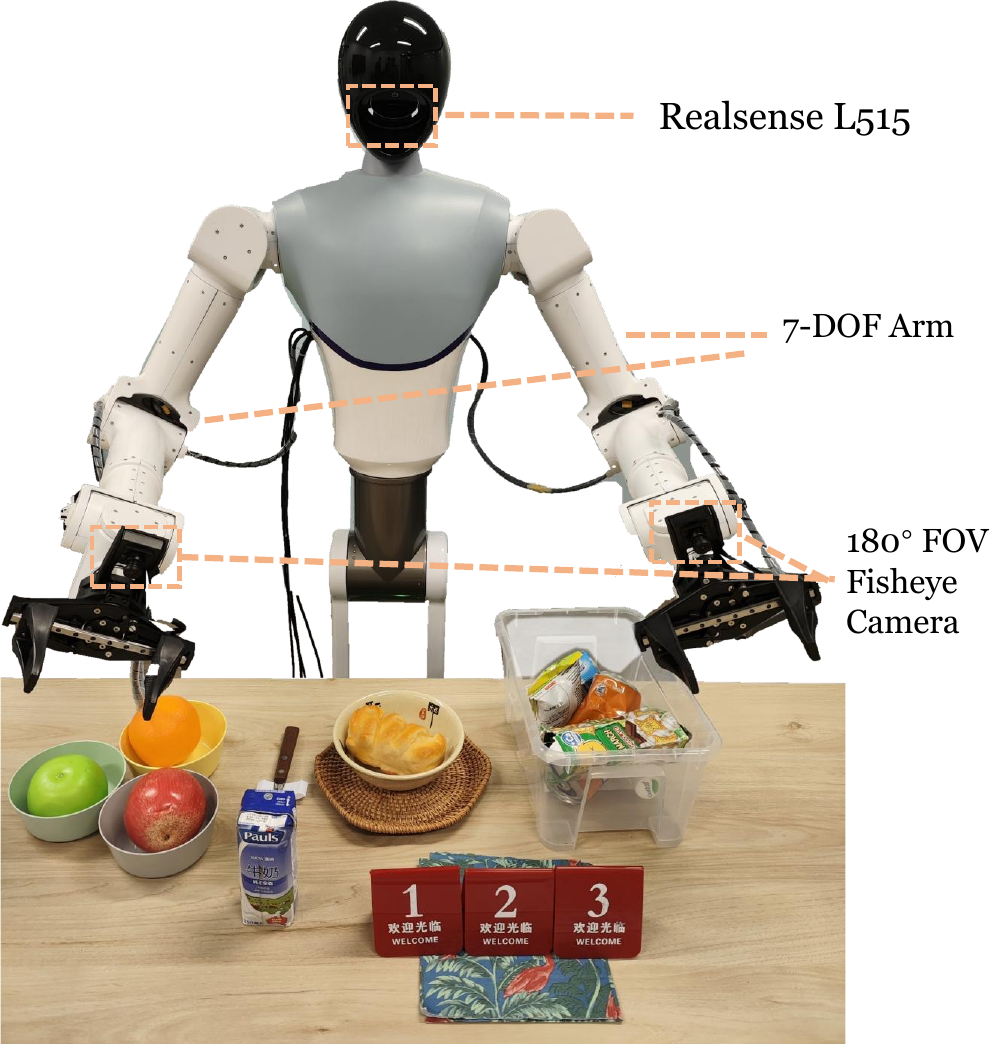}
    \caption{\textbf{Custom dual-arm robot platform used in the real-world experiments.}}
    % The platform has two 7-DoF arms, a head-mounted Intel RealSense L515 camera, and a 180-degree fisheye camera on each wrist.

    \label{fig:custom-robot-platform}
\end{wrapfigure}

Our real-world experiments use the custom dual-arm robot platform shown in Fig.~\ref{fig:custom-robot-platform}. The robot is equipped with two 7-DoF arms and parallel grippers. A head-mounted Intel RealSense L515 camera provides a global view of the workspace, while a 180-degree fisheye camera mounted on each wrist provides close-range observations during manipulation. Together, these three viewpoints capture both scene-level context and local arm--object interactions. All RGB images are resized to \(224\times224\) before being passed to the models.

The proprioceptive state has 16 dimensions, comprising seven joint values and one gripper value per arm. Joint actions are represented as offsets from the current state, whereas gripper actions are specified as absolute values. The model pads the action representation to 32 dimensions and converts its predictions back to the robot control interface for execution. Both policy configurations use the same observation and action interfaces.

\WFclear

% \subsection{Runtime scheduling}

% The action policy and progress evaluator are invoked in parallel at action replanning calls. Intermediate observation updates maintain the visual histories without invoking the grounding pipeline or the models. Model computation, grounding, and communication contribute to the achieved replanning rate; the 20~Hz robot execution frequency is not a claim of 20 model-inference calls per second. We do not report an unmeasured latency improvement.

\section{Additional Results and Analysis}\label{app:additional-results}

\subsection{Per-environment compositional results}
\label{app:compose-results}
\begin{figure*}[t]
\centering
\includegraphics[width=\textwidth]{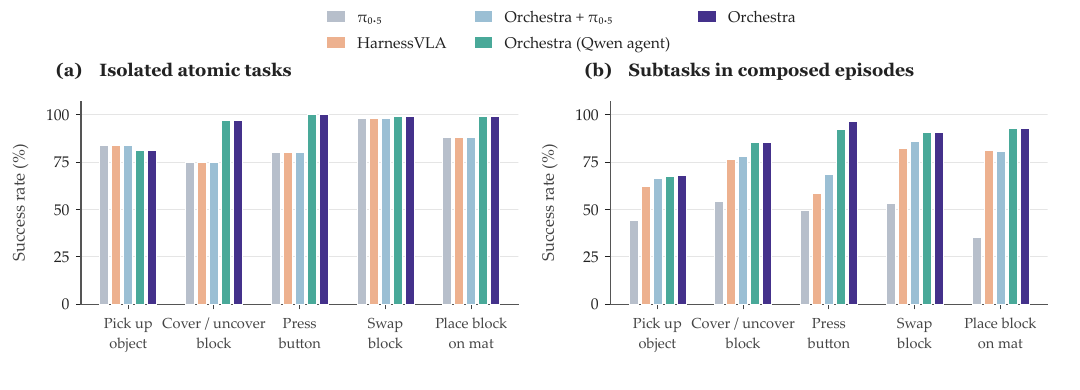}
\caption{\textbf{Per-environment success rates on RMBench-Compose.} \textbf{(a)} Atomic SR measures isolated atomic-task execution. \textbf{(b)} SSR measures success among attempted subtasks in composed episodes; unreached subtasks are excluded. Results are shown for all five evaluated configurations and all five environments. Task-level ASL is reported in Fig.~\ref{fig:results-compose}.}
\label{fig:compose-rates}
\end{figure*}

Figure~\ref{fig:compose-rates} compares isolated Atomic SR and composed-task SSR across all five evaluated configurations and environments, complementing the task-level ASL results in Fig.~\ref{fig:results-compose}. The base $\pi_{0.5}$ policy, HarnessVLA, and Orchestra with the base policy matched in Atomic SR in each of the five environments, including $98\%$ on Swap Block. On that task, adding Orchestra to the base policy raised SSR from approximately $53\%$ to $86\%$, and mean ASL increased from about one subtask to more than six.

\subsection{Ablation implementation details}

For the no-orchestration ablation in Section~\ref{sec:results-ablation}, we retrain the policy on long-horizon instructions and corresponding demonstrations while retaining short-horizon visual memory and spatial grounding. This design removes subtask decomposition while keeping the training and evaluation distributions aligned. In the visual-memory ablation, the action policy receives only the current frame from each camera. The evaluator retains its six-frame history. The spatial-grounding ablation omits task-relevant spatial conditions. The oracle-transition configuration replaces the learned completion detector with an environment-state indicator while retaining the rest of the system.

\subsection{Qualitative visualizations}
\label{app:qualitative-results}

We provide qualitative visualizations of Orchestra on
RMBench-Compose and the real-world tasks, complementing the
quantitative results in Sections~\ref{sec:results-compose}
and~\ref{sec:results-realworld}.

\subsubsection{RMBench-Compose}

\begin{figure*}[t]
    \centering
    \includegraphics[width=\textwidth]
    {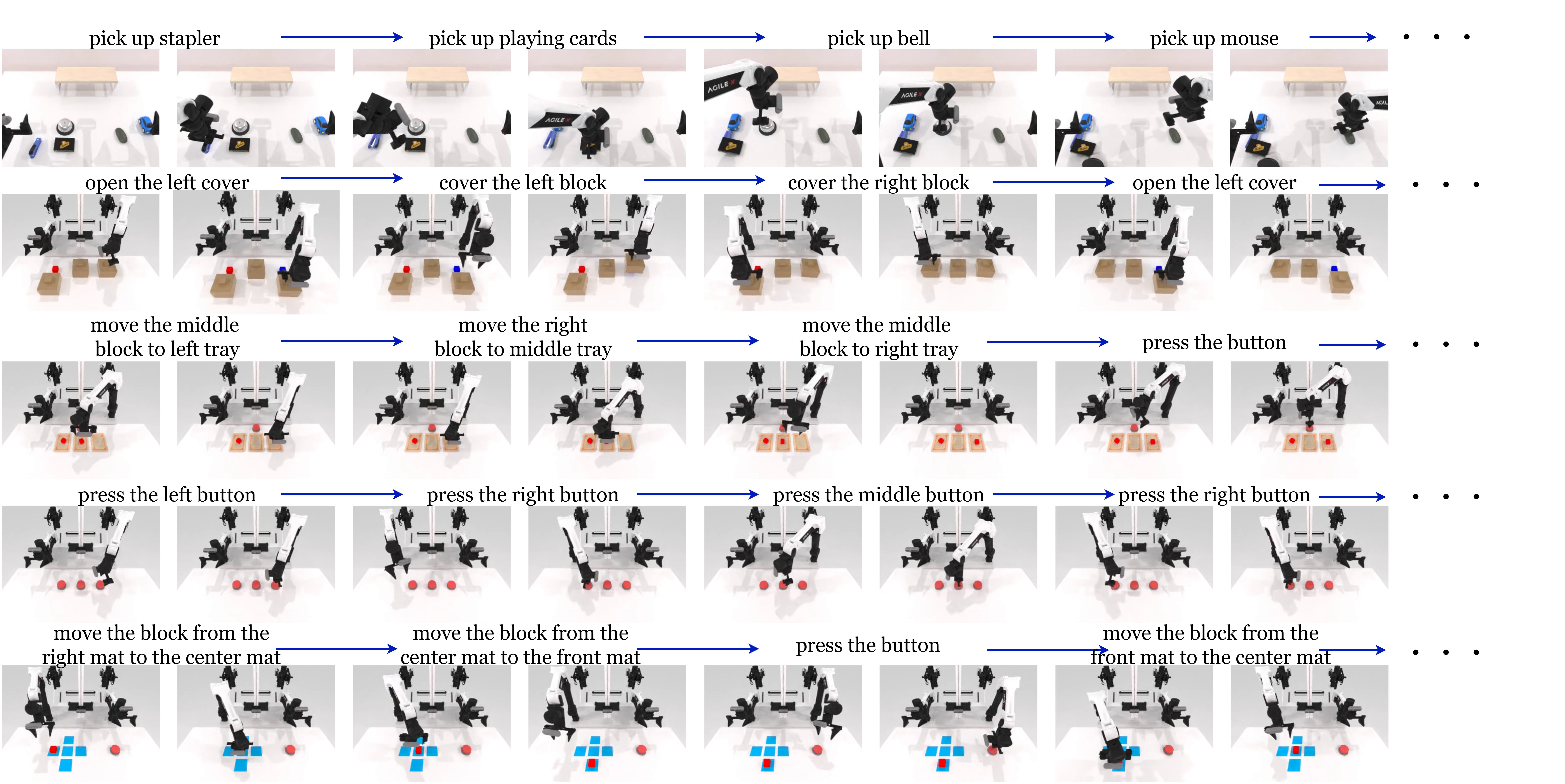}
    \caption{
    \textbf{Qualitative execution examples on RMBench-Compose.}
    From top to bottom: Pick Up Object, Cover and Uncover Block,
    Swap Block, Press Button, and Place Block on Mat.
    Each row shows selected observations from a composed
    manipulation sequence, with blue arrows indicating
    temporal progression. The examples visualize object
    manipulation and scene changes across successive subtasks;
    they do not depict every step of the full evaluation
    instructions.
    }
    \label{fig:app-compose-visualization}
\end{figure*}

Figure~\ref{fig:app-compose-visualization} presents selected
execution sequences from the five RMBench-Compose environments.
The examples illustrate how successive manipulation subtasks
change the scene, with later actions operating on the object
configurations produced by earlier actions. Frames are arranged
chronologically from left to right and show selected portions
of the composed rollouts.

\subsubsection{Real-world execution}

\begin{figure*}[p]
    \centering
    \includegraphics[
        width=\textwidth,
        height=0.95   \textheight,
        keepaspectratio
    ]{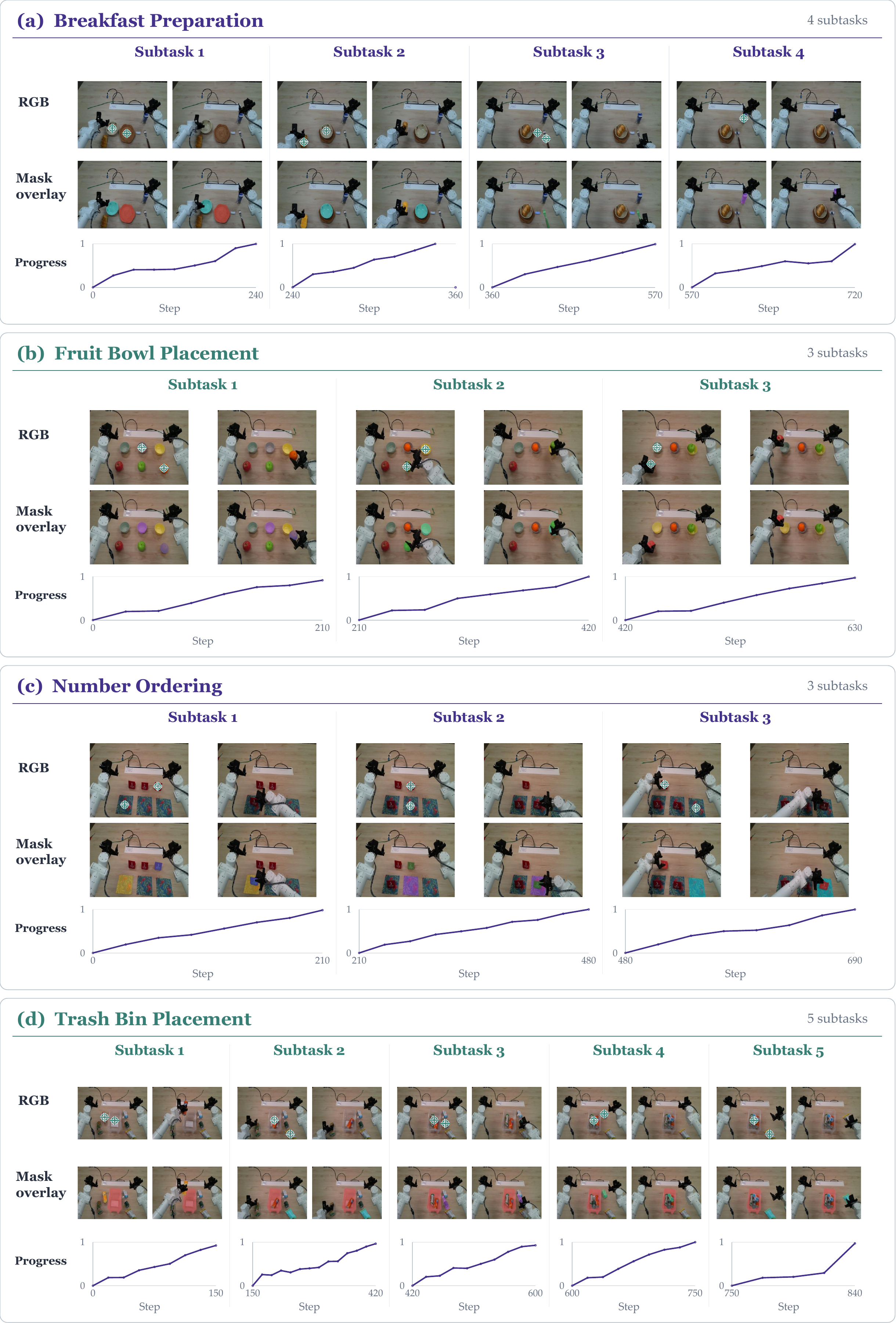}
    \caption{
    \textbf{Qualitative real-world execution with spatial
    grounding and progress visualization.}
    \textbf{(a)} Breakfast Preparation (four subtasks).
    \textbf{(b)} Fruit Bowl Placement (three subtasks).
    \textbf{(c)} Number Ordering (three subtasks).
    \textbf{(d)} Trash Bin Placement (five subtasks).
    Each subtask contains selected RGB observations with
    pointing markers, corresponding mask overlays, and a
    progress trace on a normalized scale from 0 to 1.
    Subtasks are arranged from left to right, and the
    horizontal axes of the progress plots indicate execution
    steps.
    }
    \label{fig:app-realworld-visualization}
\end{figure*}

Figure~\ref{fig:app-realworld-visualization} shows real-world
execution examples for Breakfast Preparation, Fruit Bowl
Placement, Number Ordering, and Trash Bin Placement.
For each subtask, the visualization combines selected RGB
observations, spatial mask overlays, and a within-subtask
progress trace. These views illustrate the relationship
between the active manipulation goal, its grounded entities,
and execution progress. Object pointing and segmentation are
initialized at subtask boundaries, and the initialized masks
are tracked at subsequent action-chunk replanning calls.
The plotted progress values provide a visualization of
within-subtask evolution; the deployed transition rule
uses the terminal-bin argmax decision.

\subsection{Failure case analysis}
\label{app:failure-cases}

\begin{figure*}[t]
    \centering
    \includegraphics[width=\textwidth]{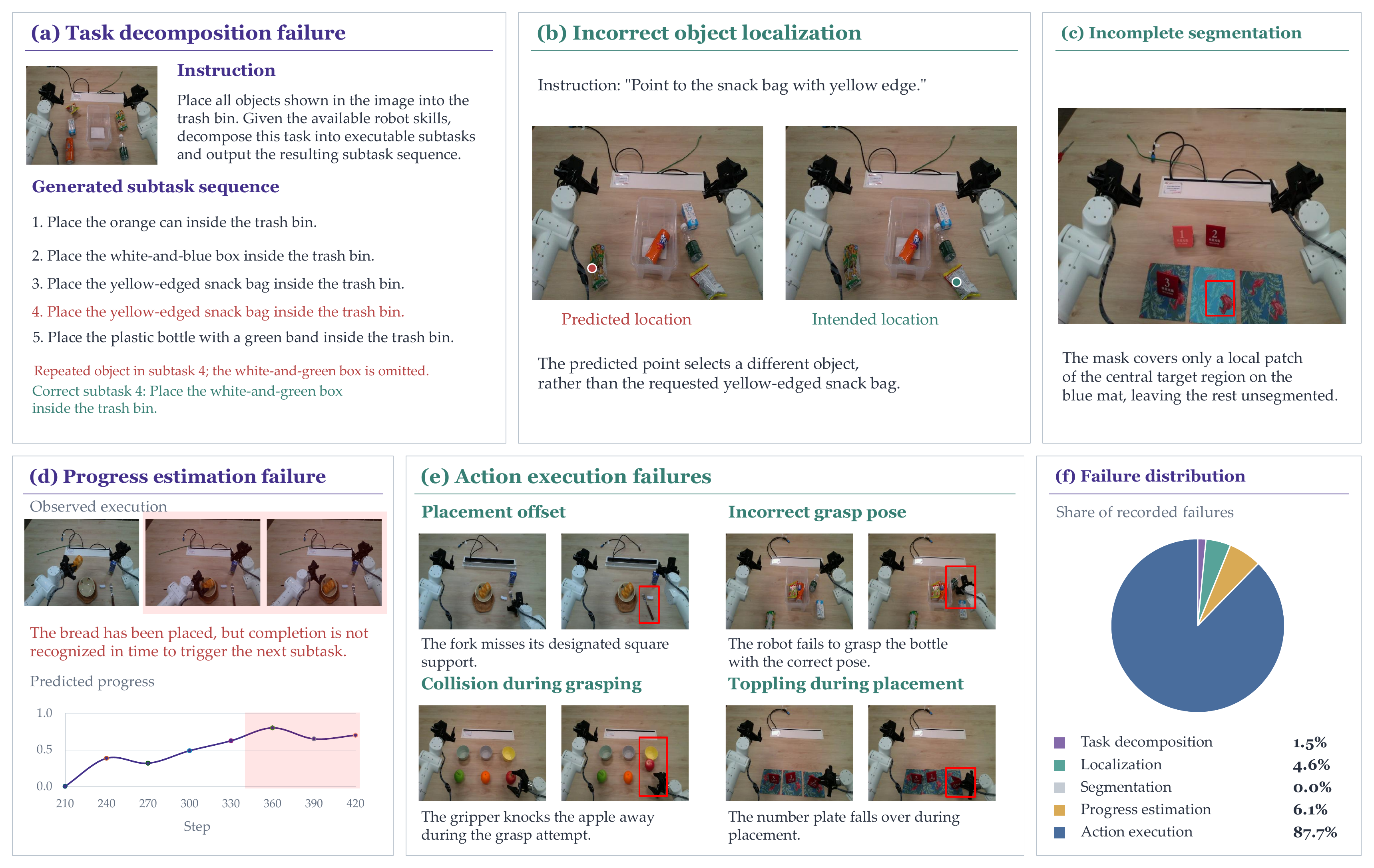}
    \caption{\textbf{Failure analysis of Orchestra in real-world manipulation.}
    \textbf{(a)} Task decomposition repeats the yellow-edged snack bag and omits the white-and-green box.
    \textbf{(b)} Object localization selects a different object from the requested snack bag.
    \textbf{(c)} Segmentation captures only part of the central target region on the blue mat.
    \textbf{(d)} The evaluator fails to recognize completion promptly after bread placement, delaying the subtask transition.
    \textbf{(e)} Action execution fails through inaccurate placement, an incorrect grasp pose, unintended contact, or unstable placement.
    \textbf{(f)} Distribution of 65 recorded failures of the complete Orchestra system across the five categories.}
    \label{fig:failure-cases}
\end{figure*}

We analyze 65 real-world failure samples from the complete Orchestra system to identify limitations at different stages of execution. These samples include both subtask-level failures and failures of the overall task. In some trials, changes in the environment necessitated a restart; these interrupted trials were excluded from the counted evaluation rollouts but retained in the failure-case analysis. Fig.~\ref{fig:failure-cases} presents errors in task decomposition, object localization, segmentation, progress estimation, and action execution, together with their shares of the recorded failures.

\paragraph{Task decomposition.}
The orchestration agent can produce individually executable subtasks that collectively fail to cover the instruction. In Fig.~\ref{fig:failure-cases}(a), the instruction requires placing all objects in the trash bin, but the generated sequence includes the yellow-edged snack bag twice and omits the white-and-green box. Executing this sequence successfully would therefore still leave the overall task incomplete. This case highlights the need to check object coverage and duplicate entity assignments when constructing a plan.

\paragraph{Object localization and segmentation.}
Spatial grounding can fail through either incorrect entity selection or incomplete region extraction. In Fig.~\ref{fig:failure-cases}(b), the localization module points to another object instead of the requested yellow-edged snack bag, providing spatial context inconsistent with the active instruction. In Fig.~\ref{fig:failure-cases}(c), the segmentation output covers only a local patch of the central target region on the blue mat. These examples show that obtaining a valid point or a nonempty mask does not guarantee semantically correct and spatially complete grounding.

\paragraph{Progress estimation.}
Physical completion and the evaluator's completion decision can diverge. In Fig.~\ref{fig:failure-cases}(d), the robot has placed the bread, but the evaluator does not recognize completion in time to trigger the next subtask. The agent advances only when the predicted progress category is the terminal bin. This failure illustrates how delayed completion detection can prevent a locally successful action from advancing the overall task.

\paragraph{Action execution.}
Fig.~\ref{fig:failure-cases}(e) illustrates limitations in grasping and placement. The robot places the fork away from its designated square support, approaches the bottle with an incorrect grasp pose, knocks the apple away during a grasp attempt, and topples the number plate during placement. These failures indicate that task orchestration and spatial conditioning do not eliminate the need for precise motion generation and stable contact handling.

\paragraph{Failure distribution and implications.}
As shown in Fig.~\ref{fig:failure-cases}(f), action execution accounts for $87.7\%$ of the 65 recorded failures, followed by progress estimation ($6.1\%$), localization ($4.6\%$), and task decomposition ($1.5\%$). Segmentation accounts for $0.0\%$ in this tally, despite the incomplete-mask example in panel~(c); observing a segmentation defect does not by itself establish that it caused a task failure. Within this evaluation, the distribution identifies action execution as the dominant remaining limitation, while the qualitative examples also motivate better plan validation, grounding checks, and completion detection.

\end{document}